\documentclass[lettersize,journal]{IEEEtran}
\usepackage{amsfonts}
\usepackage{algorithmic}
\usepackage{array}
\usepackage{textcomp}
\usepackage{stfloats}
\usepackage{verbatim}
\usepackage{graphicx}
\usepackage{comment}
\usepackage{epsfig}
\usepackage{amsmath,amssymb} % define this before the line numbering.
\usepackage{color}
\usepackage[table]{xcolor}
\usepackage{tikz}
\usepackage{multirow}
\usepackage[hang,flushmargin]{footmisc} 
\usepackage{booktabs,tabularx}
\usepackage{enumitem}
\usepackage{makecell}
\usepackage{algorithm}
\usepackage{threeparttable}
\usepackage{lipsum}
\usepackage{url}

\usepackage{caption}
\usepackage{colortbl}
\usepackage{multicol}
\usepackage{subcaption}
\usepackage{marvosym}
\usepackage[pagebackref=false, breaklinks=true, colorlinks, bookmarks=false]{hyperref}

\definecolor{linecolor}{rgb}{0.82, 0.94, 0.75}

\definecolor{lowred}{RGB}{238,18,137}

\definecolor{lowerred}{RGB}{255,110,180}

\def\ours{\textsc{{Hermes++}}}
\def\oursiccv{\textsc{{Hermes}}}

\def\BibTeX{{\rm B\kern-.05em{\sc i\kern-.025em b}\kern-.08em
    T\kern-.1667em\lower.7ex\hbox{E}\kern-.125emX}}
\usepackage{balance}
\begin{document}
\title{\textsc{{Hermes++}}: Toward a Unified Driving World Model for 3D Scene Understanding and Generation}
\author{Xin Zhou, Dingkang Liang, Xiwu Chen, Feiyang Tan, Dingyuan Zhang, \\Hengshuang Zhao, Xiang Bai\textsuperscript{\Letter}, Fellow, IEEE
\thanks{
\IEEEcompsocthanksitem X. Zhou, D. Liang, D. Zhang and X. Bai are with Huazhong University of Science and Technology. 
E-mail: (xzhou03, dkliang, xbai)@hust.edu.cn
\IEEEcompsocthanksitem X. Chen, and F. Tan are with Afari Intelligent Drive.
\IEEEcompsocthanksitem H. Zhao is with the University of Hong Kong.
\IEEEcompsocthanksitem \textsuperscript{\Letter} Corresponding author.
}}

\markboth{Journal of \LaTeX\ Class Files,~Vol.~18, No.~9, September~2020}%
{How to Use the IEEEtran \LaTeX \ Templates}

\maketitle

\begin{abstract}
Driving world models serve as a pivotal technology for autonomous driving by simulating environmental dynamics. However, existing approaches predominantly focus on future scene generation, often overlooking comprehensive 3D scene understanding. Conversely, while Large Language Models (LLMs) demonstrate impressive reasoning capabilities, they lack the capacity to predict future geometric evolution, creating a significant disparity between semantic interpretation and physical simulation. To bridge this gap, we propose \ours, a unified driving world model that integrates 3D scene understanding and future geometry prediction within a single framework. Our approach addresses the distinct requirements of these tasks through synergistic designs. First, a BEV representation consolidates multi-view spatial information into a structure compatible with LLMs. Second, we introduce LLM-enhanced world queries to facilitate knowledge transfer from the understanding branch. Third, a Current-to-Future Link is designed to bridge the temporal gap, conditioning geometric evolution on semantic context. Finally, to enforce structural integrity, we employ a Joint Geometric Optimization strategy that integrates explicit geometric constraints with implicit latent regularization to align internal representations with geometry-aware priors. Extensive evaluations on multiple benchmarks validate the effectiveness of our method. \ours~achieves strong performance, outperforming specialist approaches in both future point cloud prediction and 3D scene understanding tasks. The model and code will be publicly released at \url{https://github.com/H-EmbodVis/HERMESV2}.

\end{abstract}


\section{Introduction}
\IEEEPARstart{D}{riving} world models~\cite{chen2024end,gao2024vista,zhao2024drivedreamer,tu2026drivingworldmodel} show great potential for enhancing autonomous driving reliability by simulating environmental dynamics. These models enable vehicles to forecast risks and optimize decisions. Existing research primarily focuses on predicting scene evolution, targeting either visual appearance changes~\cite{gao2024vista,hu2023gaia} or 3D geometric deformations~\cite{zheng2025occworld,yang2024visual}. While the former captures visual texture, the latter, often represented by point clouds~\cite{hou2023query,li2023dds3d,yang2024visual,liang2025parameter}, preserves explicit geometric relationships between objects and surroundings. Maintaining accurate 3D structure is essential for downstream tasks requiring precise spatial reasoning, making it ideal for describing scene evolution.

\begin{figure*}[t]
	\begin{center}
	\includegraphics[width=0.98\linewidth]{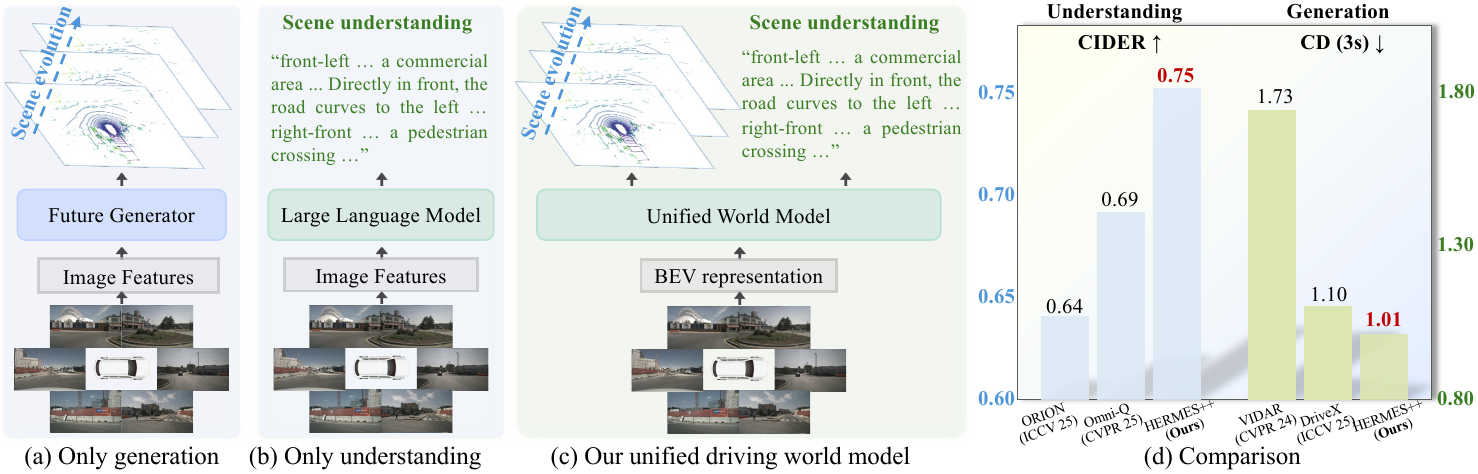}
	\end{center}
	\caption{
    (a) Previous driving world models focus on generative scene evolution prediction. (b) Large language models for driving are limited to scene understanding. (c) The proposed framework unifies 3D scene understanding and scene evolution generation with the BEV representation. (d) Quantitative comparison with dedicated specialist methods.
}
\label{fig:intro}
\end{figure*}

Despite progress in scene generation, a crucial limitation of existing methods is their limited capacity to understand 3D scenes. While capable of predicting plausible future states, they often fail to articulate the semantic context or the causal factors driving the predicted evolution. As shown in Fig.~\ref{fig:intro}(a), while driving world models excel at forecasting environmental changes, they lack the intrinsic mechanisms to answer direct queries (e.g., visual question answering, scene description). This disconnect between prediction and interpretation creates a significant capability gap, as the contextual awareness that is essential for real-world driving remains largely unaddressed by generation-centric architectures.

Moreover, recent advances in Vision-Language Models (VLMs)~\cite{liu2023visual,chen2024internvl,li2024monkey} have demonstrated remarkable capabilities in general vision tasks by leveraging world knowledge and causal reasoning from large-scale pretraining. When adapted to autonomous driving scenarios~\cite{wang2024omnidrive,sima2024drivelm,xu2024drivegpt4}, these models excel at interpreting complex driving environments, answering queries about traffic participants, generating comprehensive scene descriptions, and reasoning about spatial relationships between entities, as shown in Fig.~\ref{fig:intro}(b). For instance, OmniDrive~\cite{wang2024omnidrive} combines 3D representations with language models for visual question answering, while DriveLM~\cite{sima2024drivelm} employs graph-based reasoning for scene understanding and planning. However, these language-centric approaches prioritize understanding the current state, lacking the predictive capacity to anticipate how the scene geometry will evolve. This deficiency is critical in safety-critical scenarios where collision avoidance requires anticipating both the present context and future changes.

Motivated by the complementary strengths and limitations of these two paradigms, we propose that a world model should seamlessly integrate 3D scene understanding with accurate future geometry prediction. Constructing such a cohesive framework requires the careful consideration of two critical aspects. First, a suitable 3D representation is essential for effectively handling both textual understanding and multi-view spatial relationships. This representation must consolidate observations into a structure that preserves geometric interactions while remaining compatible with token-based language models. Second, an interaction mechanism is needed to bridge the gap between understanding and future generation. This ensures that semantic understanding guides geometric evolution and that geometric predictions ground language generation, going beyond multi-task feature sharing. Additionally, ensuring consistency in predicted scene evolutions is challenging, as supervision based solely on future observations often provides only explicit constraints, leading to structural inconsistencies.

Based on these observations and analyses, in this paper, we propose a unified driving world model that integrates understanding and generation tasks, termed \ours, as shown in Fig.~\ref{fig:intro}(c). \ours~is built upon a Bird's-Eye View (BEV) representation that naturally consolidates multi-view spatial information while ensuring compatibility with LLMs. For the linking mechanism, we introduce world queries enhanced by the LLM to transfer world knowledge from text understanding to future scene generation. These queries interact with LLM-processed BEV features via a Current-to-Future Link, ensuring that predicted scene evolution is conditioned on both geometric context and semantic reasoning.

Specifically, the BEV representation mitigates the effects of token length constraints when processing high-resolution multi-view inputs. Instead of directly converting multiple views into tokens, a BEV tokenizer consolidates them in two stages. First, a vision encoder transforms multi-view images into the BEV space using cross-attention, compressing high-dimensional inputs while preserving spatial information. Second, the BEV features are downsampled and flattened into LLM-compatible tokens. This approach reduces redundancy while maintaining geometric relationships in a consistent coordinate system. To strictly enforce geometric consistency in predicted scene evolutions, we further propose a Joint Geometric Optimization strategy. This mechanism integrates explicit geometric constraints on point clouds with implicit geometric regularization on the latent manifold. By aligning representations to geometry-aware priors, our approach ensures structural integrity throughout the generation process.

Furthermore, we introduce a knowledge transfer mechanism that bridges scene understanding with future evolution prediction. To achieve this, we directly initialize world queries from BEV features before LLM processing. These queries leverage causal attention to aggregate rich world knowledge and semantic context from text tokens. The queries then interact with LLM-encoded BEV features to generate latent representations for future timestamps via a module termed Current-to-Future Link. We further equip this link with two refinements. A Textual Injection integrates text embeddings as conditioning signals, so that semantic information directly modulates the generation process. An Ego Modulation adaptively adjusts spatial feature distributions according to future ego-motion, which decouples motion from inherent scene dynamics and ensures controllability across prediction horizons.

By conducting 3D scene understanding and future scene generation within a single framework, \ours~establishes a shared representation that seamlessly accommodates both tasks, offering a holistic perspective on driving environments. This marks a significant step toward a unified driving world model, demonstrating the feasibility of integrated driving understanding and generation. Extensive experiments validate the effectiveness of \ours~in both tasks. Notably, our method significantly reduces error by 8.2\% compared to the leading method DriveX~\cite{shi2025drivex} for challenging 3s point cloud generation. Additionally, for the understanding task, it outperforms the prior specialist baseline Omni-Q~\cite{wang2024omnidrive} by 9.2\% on the OmniDrive-nuScenes dataset~\cite{wang2024omnidrive} under the CIDEr metric.

Overall, this paper presents an early and solid exploration of a unified driving world model. By analyzing the distinct requirements of 3D scene understanding and future geometry evolution prediction, we design key components, including unified representation, world queries, and a Joint Geometric Optimization strategy. We hope this work will establish a foundation for the emerging field of interpretable and predictive autonomous driving systems. Our main contributions are summarized as follows:
\begin{itemize}
    \item We propose a unified framework that effectively integrates 3D scene understanding and future geometry prediction. By leveraging a unified representation, our method consolidates multi-view spatial information while maintaining compatibility with LLM processing.
    \item We devise a Joint Geometric Optimization strategy to enforce structural integrity in future predictions. This mechanism combines explicit geometric constraints from ground-truth point clouds with implicit geometric regularization on the latent manifold, ensuring that the predicted features align with intrinsic 3D geometry.
    \item We introduce LLM-enhanced world queries that facilitate knowledge transfer, allowing semantic reasoning derived from scene understanding to directly guide the generation of future scene evolution.
    \item We conduct extensive experiments, demonstrating that \ours~achieves strong performance across both generation and understanding, outperforming prior unified baselines and several specialist approaches. These results validate the effectiveness of the unified architecture, offering a new perspective for constructing holistic driving world models.
\end{itemize}

This paper is an extended version of our conference paper, published in ICCV 2025~\cite{zhou2025hermes}, where we make the following new contributions: \textbf{1)} We introduce a Joint Geometric Optimization strategy as an implicit geometric regularization on the latent space. The conference version relies solely on explicit point cloud constraints, whereas this strategy constructs geometry-aware representations that facilitate more accurate point cloud decoding and thereby enhance future generation performance. \textbf{2)} We further equip the Current-to-Future Link with two refinements. A Textual Injection integrates text embeddings as explicit conditioning signals, so that semantic reasoning from the language model guides the prediction of future scene evolution. An Ego Modulation adaptively adjusts spatial feature distributions, which decouples ego-motion from inherent scene dynamics and improves controllability. \textbf{3)} Through these advancements, our model achieves significant performance gains over the conference baseline. Specifically, we observe a 13.7\% reduction in generation error and consistent improvements in scene understanding metrics compared with the conference version. \textbf{4)} We have made improvements to the quality of the manuscript in various aspects. We extend the evaluations on three additional benchmarks to validate the generalization capabilities. Furthermore, we provide expanded ablation studies and in-depth discussions on the scalability of unified architectures and the intrinsic synergy between understanding and geometric evolution. These analyses not only substantiate our technical contributions but also offer insights into the potential of foundational World Models for interpretable autonomous driving.

\section{Related Work}
\label{sec:related_work}
\subsection{World Models for Driving} 

Driving world models~\cite{ha2018world} have garnered considerable attention in autonomous driving due to their ability to learn comprehensive environmental representations and predict future evolution based on action sequences. By simulating the dynamics of the driving environment, these models provide essential support for downstream tasks such as risk assessment and motion planning. Current research primarily focuses on generation tasks operating in either 2D~\cite{wang2024drivedreamer,ma2024unleashing,zheng2024doe} or 3D~\cite{min2024driveworld,ma2024cam4docc} spaces.

Most pioneering 2D world models perform video generation for driving scenarios. GAIA-1~\cite{hu2023gaia} introduced a learned simulator based on an autoregressive model. Subsequent works further leverage large-scale data~\cite{yang2024generalized,jia2023adriver,zhang2024bevworld} and advanced pre-training techniques to enhance generation quality regarding consistency~\cite{wang2024driving,gao2023magicdrive}, resolution~\cite{gao2024vista,jia2023adriver}, and controllability~\cite{zhao2024drivedreamer,wen2024panacea,li2024drivingdiffusion}. More recent approaches explore scalable DiT-based architectures~\cite{li2025uniscene,ni2025maskgwm,lu2025wovogen}, autoregressive transformers~\cite{chen2025drivinggpt,hu2024drivingworld,zhang2025epona}, and multimodal conditioning strategies~\cite{zhao2025drivedreamer4d,li2024drivingdiffusion} to improve temporal coherence. On the other hand, other studies focus on generating 3D spatial information to provide geometric representations for autonomous systems. OccWorld~\cite{zheng2025occworld} targets future occupancy generation and planning using spatial-temporal transformers, which have been adapted to other paradigms, including diffusion~\cite{wang2024occsora,gu2024dome}, rendering~\cite{agro2024uno,huang2025neural,yan2024renderworld}, and autoregressive transformers~\cite{wei2024occllama}. Some approaches propose future point cloud~\cite{zyrianov2024lidardm,weng2022s2net,khurana2023point,zhang2023learning} or depth forecasting~\cite{hassan2025gem,guo2025dist,liang2026seeing} as world models. Among these, ViDAR~\cite{yang2024visual} uses images to predict future point clouds in a self-supervised manner, while recent methods~\cite{li2025uniscene} explore geometry-aware architectures and multi-scale temporal modeling to enhance prediction accuracy.

However, existing driving world models often fail to incorporate an understanding of the driving environment. While capable of predicting future states, they lack the intrinsic ability to interpret or reason about the scenes they generate. Recent research has shifted towards unified models that combine generation and understanding within a single framework~\cite{zheng2024doe,xu2025occ,wei2024occllama,zhang2025epona,lu2025uniugp}, yet the exploration of such unified capabilities remains nascent. For example, Doe-1~\cite{zheng2024doe} explores closed-loop autonomous driving with a world model primarily focusing on single-view generation. Epona~\cite{zhang2025epona} employs an autoregressive diffusion framework to decouple temporal dynamics from visual generation for consistent long-horizon prediction and planning. FSDrive~\cite{zeng2025futuresightdrive} introduces a visual spatio-temporal Chain-of-Thought to bridge perception and planning by generating future frames with physical constraints. Despite these advances, these methods mostly operate in 2D single-view images or lack dense 3D geometric constraints intertwined with semantic reasoning.

In this paper, we propose a unified world model that understands driving scenarios and generates future geometric scene evolution, establishing a holistic framework for interpretable and predictive autonomous driving.

\subsection{Large Language Models for Driving}

Large Language Models (LLMs) and Vision-Language Models (VLMs) have achieved significant success by leveraging extensive world knowledge and causal reasoning capabilities derived from large-scale pretraining~\cite{wu2024visionllm,zhang2025psalm,dong2023dreamllm,cho2025language,fu2026minddrive}. In the realm of autonomous driving, these models effectively bridge the gap between raw sensory data and semantic understanding, enabling the interpretation of complex traffic scenarios, reasoning about agent behaviors, and generating natural language explanations. Such capabilities are crucial for developing reliable autonomous systems capable of handling diverse driving situations.

Recent research has adapted LLMs and VLMs to various driving tasks. For scene understanding, DriveGPT4~\cite{xu2024drivegpt4} employs a VLM to generate driving commands alongside natural language justifications based on front-view observations. DriveLM~\cite{sima2024drivelm} introduces scene graphs to facilitate structured reasoning and end-to-end driving via graph-based visual question answering. Similarly, OmniDrive~\cite{wang2024omnidrive} integrates 3D spatial representations with VLMs using a Q-Former, establishing a comprehensive benchmark for multi-task driving comprehension. To enhance spatial-temporal modeling, ELM~\cite{zhou2025embodied} proposes pre-training strategies tailored specifically for embodied scenarios. Beyond perception and reasoning, Vision-Language-Action (VLA) models have emerged to directly link perception with control~\cite{cui2025drivemlm,hwang2024emma, tian2025drivevlm,shao2024lmdrive}. For example, ORION~\cite{fu2025orion} performs a differentiable connection between reasoning and action space. Despite these advances, existing methods primarily rely on the LLM to understand the current state, often lacking the capacity to predict the future geometric evolution of the surrounding environment.

In this paper, we bridge this gap by enabling LLMs to comprehend the present driving scenario and predict its future evolution. Rather than treating these as isolated tasks, we establish dedicated mechanisms that allow semantic reasoning from language understanding to guide geometric prediction. This design empowers the model to leverage world knowledge for generating structurally coherent future scenes, creating a framework that seamlessly integrates scene comprehension with accurate prediction.

\section{Preliminaries}
This section briefly reviews driving world models and the Bird's-Eye View representation as preliminaries.

\textbf{Driving world models} aim to learn a general representation of the driving environment by forecasting the future dynamics of a scene~\cite{ha2018world,wang2024drivedreamer,zhou2025hermes,zheng2025occworld,yang2024visual}. The core objective is to predict future states based on current observations and planned actions, enabling the model to capture the underlying data distribution of real-world driving scenarios.

Formally, given an observation $\mathcal{O}_{t}$ at time $t$ and an action $\mathcal{A}_{t}$, a driving world model predicts the subsequent observation $\mathcal{O}_{t+1}$. This process typically involves three components:
\begin{equation} 
\mathcal{Z}_t = E \left ( \mathcal{O}_{t}\right), ~ \mathcal{Z}_{t+1} = M\left ( \mathcal{Z}_{t}, \mathcal{A}_{t}\right), ~ \hat{\mathcal{O}}_{t+1} = D \left ( \mathcal{Z}_{t+1}\right), 
\end{equation}
where $E:\mathcal{O}\to \mathcal{Z}$ is an encoder that maps observations to latent representations, $M:\mathcal{Z}\times\mathcal{A}\to \mathcal{Z}$ is the predicting model that transitions the latent state forward in time conditioned on an action, and $D:\mathcal{Z}\to \mathcal{O}$ is a decoder that reconstructs the observation from the predicted latent state. The latent space $\mathcal{Z}$ serves as a compact representation that captures essential scene information while filtering out irrelevant details. While $\mathcal{O}_{t}$ can vary across modalities (e.g., RGB images, LiDAR), this work focuses on multi-view images as input and point clouds as output, leveraging the latter's ability to preserve 3D geometric structures, which are essential for spatial reasoning. Two further considerations support this choice of output modality. First, point clouds are directly measured by the LiDAR sensor and require no additional annotation, so the geometric supervision scales easily to larger amounts of data, whereas 3D occupancy must be constructed by aggregating sequential scans and, in its semantic form, additionally requires per-voxel categories. Second, point clouds can be decoded from the predicted representation by a lightweight differentiable renderer, whereas generating plausible future images relies on the strong priors of large pre-trained generative models, which would obscure the contribution of the unified representation studied in this work. We further note that this choice is not a restriction of the architecture, since the world queries and the Current-to-Future Link (Sec.~\ref{sec:unification}) operate on latent features and are agnostic to the output modality.

\textbf{Bird's-Eye View (BEV)} has emerged as a foundational spatial representation for autonomous driving, offering a natural coordinate system for multi-view fusion and spatial reasoning~\cite{zhang2024bevworld,li2023bevdepth,liu2023bevfusion,li2023delving}. Unlike perspective views, which are prone to occlusion and scale ambiguity, BEV preserves geometric relationships in a top-down logical space.

Given multi-view images $\{\mathcal{I}_i\}_{i=1}^{N}$ from $N$ multi-view cameras, the BEV representation is defined as a feature map $\mathbf{F}_{\text{BEV}}\in \mathbb{R}^{H\times W \times C}$, where $H \times W$ represents the spatial resolution and $C$ denotes the feature dimension. The transformation from perspective view to 3D BEV space requires lifting image features to 3D spatial locations. Following modern approaches like the BEVFormer series~\cite{li2022bevformer,yang2023bevformer}, we employ learnable grid queries $\mathbf{Q}_{\text{BEV}} \in \mathbb{R}^{H \times W \times C}$ positioned at predefined grid locations in BEV space. For each query at spatial location $(x,y)$, the corresponding BEV feature is computed through spatial cross-attention:
\begin{equation}
\mathbf{B}(x, y) = \sum_{i=1}^{N} \sum_{z \in \mathcal{H}} \text{DA}\left(\mathbf{Q}(x, y), \mathbf{F}_i, \mathcal{P}_i(x, y, z)\right), 
\end{equation}
where $\mathrm{DA}(\cdot)$ denotes multi-scale deformable cross-attention that aggregates features from the $i$-th camera feature map $\mathbf{F}_i$ around the projected reference point $\pi_i(x,y,z)$ with learned sampling offsets and attention weights.
$\mathcal{H}$ is a set of predefined height anchors, and $\pi_i(\cdot)$ maps a 3D location to the image plane using camera intrinsics and extrinsics.

By encoding geometry, the BEV representation effectively unifies visual semantics with spatial structure. Its natural integration of visual semantics and geometric structure makes BEV well-suited for both scene understanding and generation. We thus leverage it as the core substrate to bridge scene understanding and future evolution prediction within a shared geometric space.

\section{Method}
\begin{figure*}[t]
	\begin{center}
	\includegraphics[width=0.97\linewidth]{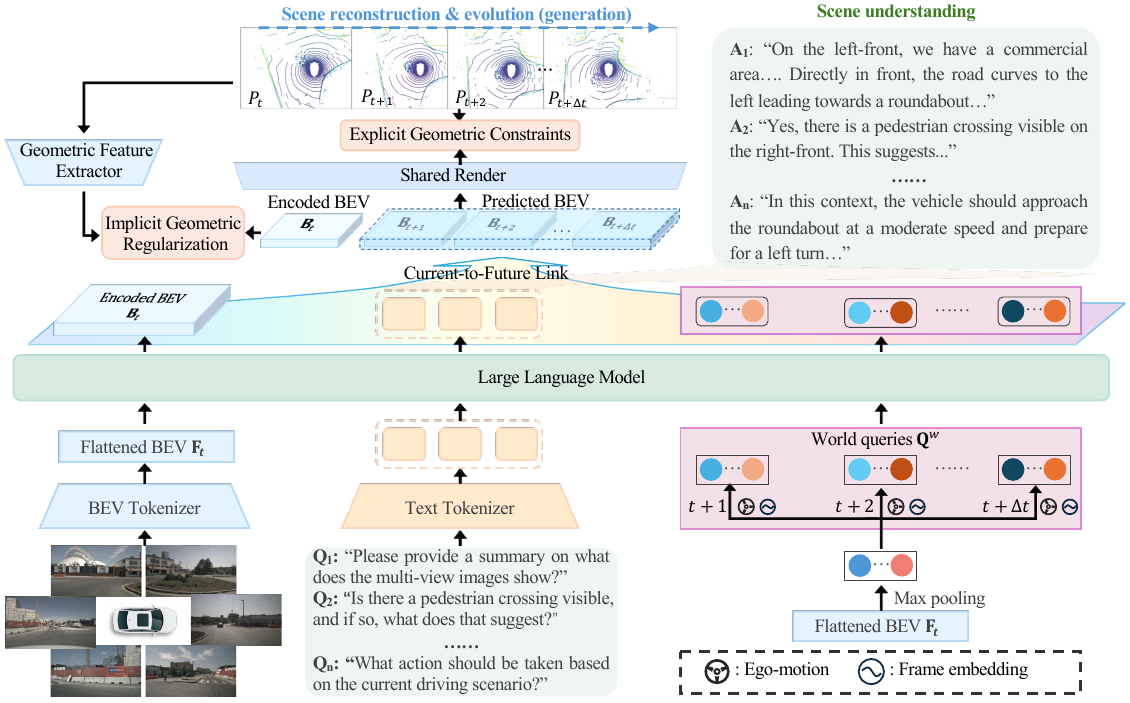}
	\end{center}
	\caption{Pipeline of \ours. Flattened BEV tokens, instructions, and world queries are input to the LLM to generate text and semantic contexts. The Current-to-Future Link propagates the encoded BEV to future states, conditioned on both textual semantics and world queries. The shared Render then predicts the evolution of the point cloud. During training, a Joint Geometric Optimization strategy ensures structural integrity by combining explicit geometric constraints on rendered outputs with implicit geometric regularization on latent features using a frozen geometry extractor.}
\label{fig:pipeline}
\end{figure*}

Fig.~\ref{fig:pipeline} illustrates the overall framework of \ours, which seamlessly integrates language-based reasoning with geometric generation. The pipeline begins by transforming multi-view images into the BEV representation, which is subsequently compressed into visual tokens compatible with the large language model. These tokens, concatenated with user instructions and learnable world queries, are processed by the LLM to generate textual responses while aggregating semantic context into the queries. Following this, the Current-to-Future Link propagates the encoded BEV features to future timestamps, conditioned on the enriched queries, text embeddings, and ego-motion. Finally, a shared Render reconstructs point clouds from the predicted features. To strictly enforce structural integrity, we employ a Joint Geometric Optimization strategy that integrates explicit geometric constraints on reconstructed point clouds with implicit geometric regularization on the latent manifold. Through this unified pipeline, \ours~effectively bridges the gap between perception and prediction, leveraging world knowledge to directly guide future scene evolution. Compared with the conference version~\cite{zhou2025hermes}, the visual tokenizer and Render in Sec.~\ref{sec:tokenizer} and the world queries in Sec.~\ref{sec:unification} are inherited. The Joint Geometric Optimization strategy in Sec.~\ref{sec:joint_supervision} is newly introduced, together with the Textual Injection and Ego Modulation that refine the Current-to-Future Link in Sec.~\ref{sec:method_link}, and the training objectives in Sec.~\ref{sec:training} are extended accordingly.

\subsection{Visual Tokenizer and BEV-to-Point Render} \label{sec:tokenizer}
To address the spatial discontinuity inherent in multi-view inputs and the token constraints of LLMs, we adopt the BEV representation, which naturally preserves geometric structural integrity while enabling efficient compression for language interaction. Specifically, a visual tokenizer is used for feature encoding and a differentiable Render for geometric decoding.

For feature encoding, we employ a BEV-based visual tokenizer to preserve geometric spatial relationships across views while providing semantically rich inputs for the LLM. Given multi-view images $\{I_t^i\}_{i=1}^{N}$ at time $t$ from $N$ cameras, the tokenizer operates in two stages. First, we encode the multi-view images using a vision encoder~\cite{cherti2023reproducible,liu2022convnet} to extract multi-scale perspective features. These features are then transformed into a BEV representation $\mathbf{F}^{\text{bev}}_t \in \mathbb{R}^{w \times h \times c}$ via spatial cross-attention~\cite{yang2023bevformer}. This process effectively captures both semantic and geometric information in the BEV space, where $w$ and $h$ denote the spatial dimensions of the BEV grid, and $c$ represents the feature channels. Because the raw BEV features contain a large number of tokens that exceed the length limits of most LLMs~\cite{zhang2024make}, we apply a downsampling module composed of strided convolutions and pooling. This reduces the spatial resolution by a factor of 4 while expanding the channel dimension to preserve information, yielding $\mathbf{F}^{\text{down}}_t \in \mathbb{R}^{\frac{w}{4} \times \frac{h}{4} \times 4c}$. Finally, we flatten and project the compressed feature through a linear layer for LLM processing, following established practices in vision-language models~\cite{liu2023visual,chen2024far}:
\begin{equation}\label{eq:bev_proj}
\mathbf{F}_t = \phi(\text{Flatten}(\mathbf{F}^{\text{down}}_t)) \in \mathbb{R}^{L_{\text{BEV}} \times C},
\end{equation}
where $L_{\text{BEV}}=\frac{w}{4} \times \frac{h}{4}$ is the number of BEV tokens, $C$ is the hidden dimension of the LLM, and $\text{Flatten}(\cdot)$ reshapes the spatial dimensions into token sequences. This tokenization balances spatial detail preservation with computational efficiency.

To complement this encoding process and recover 3D geometry from BEV spaces, we introduce a differentiable BEV-to-Point Render $\mathcal{R}$. This module operates on either the compressed BEV feature $\mathbf{F}^{\text{down}}_t$ or the LLM-processed BEV feature $\mathbf{B}_t$ (after out-projection from the LLM dimension). To generate 3D geometry from 2D BEV features, we first upsample the input to its original spatial resolution using nearest neighbor interpolation and convolutional layers. As BEV features lack height information, we expand them into a volumetric representation by reshaping along the height dimension. A series of 3D convolutional layers then refines this representation to produce $\hat{\mathbf{V}}_t \in \mathbb{R}^{w \times h \times z \times c'}$, where $z$ denotes the number of discrete height levels, and $c'$ is the output dimension encoding the 3D scene structure.

Following established neural rendering techniques~\cite{yang2024unipad,zhu2023ponderv2,wang2021neus}, we model the scene geometry as an implicit signed distance function (SDF) field. For each LiDAR ray $\mathbf{r}_k$ emitted from the center $\mathbf{o}$ along direction $\mathbf{t}_k$, we discretize it into $n$ sample points $\mathbf{p}_i = \mathbf{o} + d_i \mathbf{t}_k$ with increasing depths $0 \le d_1 < d_2 < \cdots < d_n$. For each point, we retrieve the local feature $\mathbf{f}_i$ from $\hat{\mathbf{V}}_t$ via trilinear interpolation and predict the SDF value $s_i$ using a shallow MLP $\phi_{\mathrm{SDF}}$, i.e., $s_i=\phi_{\mathrm{SDF}}(\mathbf{p}_i,\mathbf{f}_i)$. With the predicted SDF values, the rendered depth for ray $\mathbf{r}_k$ is computed as a weighted sum:
\begin{equation} 
\tilde{d}(\mathbf{r}_k) = \sum_{i=1}^n w_i d_i, \quad w_i = T_i \alpha_i,
\end{equation}
where $w_i$ represents the weight derived from transmittance $T_i = \prod_{j=1}^{i-1}(1 - \alpha_j)$ and opacity $\alpha_i$. The opacity is computed from the SDF gradients as:
\begin{equation} 
\alpha_i = \max\left(\frac{\sigma_\tau(s_i) - \sigma_\tau(s_{i+1})}{\sigma_\tau(s_i)}, 0\right),
\end{equation}
where $\sigma_\tau(x)=(1+e^{-\tau x})^{-1}$ is the sigmoid with a learnable parameter $\tau$. The final point cloud $P_t = \{\mathbf{p}_k\}_{k=1}^K$ is obtained by converting the rendered depths back to 3D coordinates.

\subsection{Unification of Understanding and Generation}
\label{sec:unification}
This subsection details the unification mechanism that seamlessly integrates scene understanding and future evolution prediction within \ours. The central challenge lies in establishing knowledge transfer from language-based reasoning to geometric prediction, thereby enabling the world knowledge acquired during understanding to guide future generation. We address this through two key designs: world queries that aggregate semantic information, and a Current-to-Future Link that leverages both geometric and textual conditioning for temporal evolution modeling.

\subsubsection{\textbf{Language-based Scene Understanding}}

The LLM serves as the central reasoning engine, processing multi-view observations encoded as BEV features $\mathbf{F}_t$ (Eq.~\ref{eq:bev_proj}) and responding to user instructions about the driving scenario. User instructions are tokenized into text tokens $\mathbf{T} \in \mathbb{R}^{L_{\text{text}} \times C}$ using the LLM's vocabulary, where $L_{\text{text}}$ is the text length. These sequences are concatenated and processed by the LLM, enabling the model to interpret the driving scene through next-token prediction. This process not only serves the immediate query-answering task but also enriches the internal representations with semantic knowledge essential for future prediction.

\subsubsection{\textbf{World Queries for Knowledge Transfer}}
To enable knowledge transfer from understanding to generation, we introduce world queries that interact directly with the LLM's processing pipeline and act as temporal semantic carriers. Unlike conventional approaches that employ separate branches for task-specific processing, our world queries are injected into the input sequence, acting as latent placeholders that aggregate semantic information from both visual and textual contexts.

Specifically, we initialize $\Delta t$ groups of world queries, with each group containing $n$ tokens corresponding to a distinct future time step. This initialization leverages adaptive max pooling over the compressed BEV features $\mathbf{F}^{\text{down}}_t$ to extract salient spatial information, yielding the base spatial queries $\mathbf{Q} \in \mathbb{R}^{n \times 4c}$, where $4c$ represents the channel dimension. To enable controllable future generation conditioned on ego-vehicle trajectories, we process the ego-motion parameters for each future frame using an MLP to generate high-dimensional motion embeddings $\mathbf{e}_{t+i} \in \mathbb{R}^{1 \times 4c}$. These embeddings are then broadcasted and added to the base spatial queries. Furthermore, learnable frame embeddings $\mathbf{FE} \in \mathbb{R}^{\Delta t \times 4c}$ are introduced to encode the temporal order. The final world queries $\mathbf{Q}^w \in \mathbb{R}^{(\Delta t \times n) \times C}$ are constructed by concatenating these temporally conditioned representations and projecting them to the hidden dimension of the language model using a shared linear layer $\phi$:
\begin{equation}
\mathbf{Q}^w = \phi\left(\text{Concat}_{i=1}^{\Delta t}(\mathbf{Q} \oplus \mathbf{e}_{t+i}) \oplus \mathbf{FE}\right),
\end{equation}
where $\oplus$ denotes element-wise addition with broadcasting across the spatial query and temporal dimensions accordingly.

We argue that our world queries facilitate knowledge transfer in two ways. First, the causal attention mechanism allows world queries to access all preceding tokens, effectively aggregating specific semantic context from current BEV observations and textual instructions. Second, the language model contains rich intrinsic world knowledge during large-scale pre-training. By processing queries, the model infuses them with world knowledge and causal priors. Thus, the resulting enriched queries $\mathbf{Q}^w_{\epsilon}$ encapsulate both context-specific details and generalized world knowledge, serving as priors for the subsequent geometric forecasting task.

\subsubsection{\textbf{Current-to-Future Link}}
\label{sec:method_link}
While the world queries capture semantic information from the understanding process, they provide only a sparse representation of future scenes with $n$ queries per time point. To generate dense future BEV features, we introduce a Current-to-Future Link that propagates spatial information from the current encoded BEV $\mathbf{B}_t$ to future time points, conditioned on both world queries and text embeddings. We further refine this link with two designs, a Textual Injection and an Ego Modulation, which condition the propagation on semantic context and on the planned ego-motion, respectively.

The Textual Injection incorporates semantic context into the generation process. Specifically, we extract text embeddings $\hat{\mathbf{T}} \in \mathbb{R}^{k \times C}$ from the LLM-processed text tokens via average pooling and linear projection, where $k$ is the number of pooled tokens and $C$ is the LLM dimension. For each future time step $i \in \{1, \ldots, \Delta t\}$, let $\mathbf{Q}^w_{\epsilon,i}$ denote the corresponding world queries. The link consists of several stacked blocks, each containing cross-attention, self-attention, and feed-forward layers. The cross-attention layer aggregates information from both world queries and text:
\begin{equation}
\mathbf{X}^{(l)}_{\text{cross}} = \mathbf{X}^{(l)} + \text{CrossAttn}(\text{LN}(\mathbf{X}^{(l)}), [\mathbf{Q}^w_{\epsilon,i}; \hat{\mathbf{T}}]),
\end{equation}
where $\mathbf{X}^{(l)}$ represents the input features (initialized as $\mathbf{B}_t$), and the concatenation $[\mathbf{Q}^w_{\epsilon,i}; \hat{\mathbf{T}}]$ serves as the Key and Value. This formulation enables the model to jointly attend to geometric context from world queries and semantic guidance from text, explicitly directing the spatial evolution.

To ensure the predicted scene aligns with planned trajectories, we further apply an Ego Modulation (EM) that adapts feature distributions based on future ego-motion. Specifically, the ego-motion for time point $t+i$ is encoded through an MLP with a Tanh activation to produce modulation parameters $\gamma$ and $\beta$ for self-attention and feed-forward layers:
\begin{equation}
\text{EM}(\mathbf{x}) = (\gamma+1) \odot \text{LN}(\mathbf{x}) + \beta,
\end{equation}
where $\odot$ denotes element-wise multiplication, and the modulation vectors are zero‑initialized at the initial period of training to maintain stability. This modulation adaptively adjusts spatial representations based on driving maneuvers. Notably, EM is applied only to self-attention and feed-forward branches, preserving the cross-attention's focus on semantic aggregation.

Finally, the processed features are reshaped and upsampled to future BEV representations $\{\mathbf{B}_{t+i}\}_{i=1}^{\Delta t}$, which are subsequently used by the shared Render to generate corresponding future point cloud evolutions $\{P_{t+i}\}_{i=1}^{\Delta t}$.

The Current-to-Future Link thus serves as a bridge between understanding and generation, enabling world knowledge from language reasoning to inform geometric prediction while maintaining controllability through ego-motion conditioning. This architecture enables the model to generate future scenes that are both semantically consistent with the current context and geometrically aligned with the vehicle's behavior.

\subsection{Joint Geometric Optimization Strategy}\label{sec:joint_supervision}
We observe that relying solely on rendering-based supervision often results in structural ambiguity in the latent representation, as it fails to capture intricate geometric structures and foreground objects. To address this, we propose a Joint Geometric Optimization strategy that imposes constraints on both the observational and latent levels. This mechanism integrates explicit geometric constraints derived from ground-truth point clouds with implicit regularization to internal representations.

\textbf{Explicit Geometric Constraints.} 
At the observational level, we supervise the generated point clouds to ensure geometric integrity. We employ a simple $L_1$ loss on the rendered depths to minimize the discrepancy between the predicted and ground-truth geometry. The rendering loss is formulated as:
\begin{equation}\label{eq:depthloss}
\mathcal{L}_{\text{render}} = \sum_{i=0}^{\Delta t} \lambda_{i} \frac{1}{N_{i}} \sum_{k=1}^{N_{i}} \left | d(\mathbf{r}_k) - \tilde{d}(\mathbf{r}_k) \right |,
\end{equation}
where $d(\mathbf{r}_k)$ and $\tilde{d}(\mathbf{r}_k)$ denote the ground-truth and predicted depths for ray $\mathbf{r}_k$, $\lambda_{i}$ is the loss weight for frame $t+i$, and $N_{i}$ is the number of rays. This explicit constraint ensures that the decoded point clouds align with the physical measurements.

\textbf{Implicit Geometric Regularization.} 
Complementary to explicit constraints, we introduce an implicit regularization for the latent manifold. This strategy aligns the predicted features with geometry-aware representations, thereby guiding the model to encode intrinsic 3D structures. Specifically, this process leverages a geometric feature extractor to derive latent priors, which are then enforced on the predicted features via two complementary alignment objectives.

To obtain the geometry-aware priors, we utilize a self-supervised point cloud reconstruction network as the geometric feature extractor. This network comprises a sparse 3D convolutional encoder paired with the differentiable Render $\mathcal{R}$. During a pre-training stage, we voxelize the ground-truth point cloud $P_t$, extract features using the extractor, and subsequently render them to reconstruct the input point cloud. This self-supervision enables the capture of spatially meaningful representations. In the main training phase, the Geometric Feature Extractor functions as a frozen encoder to produce the target geometry-aware features $\mathbf{V}_t \in \mathbb{R}^{w \times h \times z \times c'}$.

We align the predicted volumetric features $\hat{\mathbf{V}}_t$ with the frozen geometry-aware features $\mathbf{V}_t$ using two objectives that capture both local correspondence and global structural patterns. First, to enforce element-wise consistency, we employ a cosine similarity loss:

\begin{equation}
\label{eq:cos_loss}
\mathcal{L}_{\text{cos}} = 1 - \frac{1}{whz} \sum_{i,j,k} \frac{\hat{\mathbf{V}}_t(i,j,k) \cdot \mathbf{V}_t(i,j,k)}{\|\hat{\mathbf{V}}_t(i,j,k)\|_2 \|\mathbf{V}_t(i,j,k)\|_2},
\end{equation}
where $(i,j,k)$ indexes the voxel grid. Second, we introduce a Gram loss to measure feature correlations for global consistency. We pool the features along orthogonal spatial axes to obtain projected maps $\mathbf{V}^{HW}_t$, $\mathbf{V}^{HZ}_t$, and $\mathbf{V}^{WZ}_t$. For a perspective $d \in \{HW, HZ, WZ\}$, the spatial Gram matrix is formulated as:
\begin{equation} \label{eq:gram-matrix}
\mathbf{G}^d_t = \mathbf{V}^d_t {\mathbf{V}^d_t}^T \in \mathbb{R}^{N_d \times N_d},
\end{equation}
where $N_d$ is the spatial tokens length in perspective $d$. This matrix captures pairwise correlations across spatial locations. The Gram loss minimizes the discrepancy between the Gram matrices of the predicted and geometry-aware features:
\begin{equation}
\label{eq:gram_loss}
\mathcal{L}_{\text{gram}} = \frac{1}{3} \sum_{d} \|\mathbf{G}^d_t - \hat{\mathbf{G}}^d_t\|_F^2, ~ d \in \{HW, HZ, WZ\},
\end{equation}
where $\hat{\mathbf{G}}^d_t$ and $\mathbf{G}^d_t$ are computed from
$\hat{\mathbf{V}}_t$ and $\mathbf{V}_t$, respectively. $\|\cdot\|_F$ denotes the Frobenius norm. 

Since the geometry extractor is used only for training-time regularization and is discarded at inference, it introduces no additional inference-time cost.

\subsection{Training Objectives}
\label{sec:training}
The training of \ours~is governed by a composite objective function that jointly optimizes language understanding, geometric rendering, and structural alignment.

To enable scene understanding, we employ next-token prediction following the standard autoregressive language modeling objective:
\begin{equation}
\mathcal{L}_{\text{lang}} = -\sum_{i=1}^{L_{\text{text}}} \log P\left ( \mathbf{T}_{i} | \mathbf{F}_{t}, \mathbf{T}_{1}, \ldots, \mathbf{T}_{i-1}; \boldsymbol{\Theta} \right ),
\end{equation}
where $P(\cdot | \cdot)$ represents the conditional probability modeled by the LLM parameters $\boldsymbol{\Theta}$, $\mathbf{F}_{t}$ is the flattened BEV feature, and $\mathbf{T}_{i}$ denotes the $i$-th text token. To supervise future point cloud generation, we utilize the proposed Joint Geometric Optimization strategy. This objective integrates the explicit geometric constraints (Eq.~\ref{eq:depthloss}) with the implicit geometric regularization (Eq.~\ref{eq:cos_loss} and Eq.~\ref{eq:gram_loss}):
\begin{equation}
\mathcal{L}_{\text{gen}} = 10\mathcal{L}_{\text{render}} + \mathcal{L}_{\text{cos}} + \mathcal{L}_{\text{gram}}.
\end{equation}

Finally, the overall objective is the summation of the understanding and generation losses:
\begin{equation}
\mathcal{L}_{\text{total}} = \mathcal{L}_{\text{lang}} + \mathcal{L}_{\text{gen}}.
\end{equation}

\section{Experimental Setup}
\subsection{Datasets and Evaluation Metric}

We conduct comprehensive experiments on four datasets, and additionally use NuInteract for vision-language alignment.

\textbf{NuScenes Dataset}~\cite{caesar2020nuscenes} serves as the primary benchmark for geometric representation learning and unified training. In our experiments, we utilize the multi-view images as inputs and synchronized point clouds as ground truth. Following prior work~\cite{yang2024visual,zhou2025hermes}, we evaluate the consistency of predicted future scenes using the bidirectional Chamfer Distance (CD). The evaluation is restricted to a Region of Interest (ROI) defined by $x, y \in [-51.2\text{m}, 51.2\text{m}]$ and $z \in [-3\text{m}, 5\text{m}]$.

\textbf{OmniDrive-nuScenes Dataset}~\cite{wang2024omnidrive} is utilized for the refinement phase and unified instruction tuning. It enriches the original NuScenes dataset with high-quality scene descriptions and visual question-answering (QA) pairs, requiring the model to reason about complex object interactions and traffic contexts. To assess scene understanding capabilities, we evaluate the quality of textual responses on the validation set. We report CIDEr~\cite{vedantam2015cider} for consensus, METEOR~\cite{banerjee2005meteor} for semantic alignment, and ROUGE-L~\cite{lin2004rouge} for structural similarity, following the official evaluation.

\textbf{NuScenes-QA Dataset}~\cite{qian2024nuscenes} is a large-scale visual question answering benchmark, comprising $\sim$460k QA pairs. It leverages 3D detection annotations to construct scene graphs, evaluating the model's ability to interpret complex driving environments and reason about spatial relationships. We report the standard top-1 accuracy to quantify performance on this multi-view visual question-answering task.

\textbf{DriveLM Dataset}~\cite{sima2024drivelm} introduces Graph Visual Question Answering to mimic human-like reasoning in driving. Constructed on selected NuScenes keyframes, it interconnects perception, prediction, and planning questions via logical dependencies. The rich logical chain annotations assess the depth of the reasoning and its capability to align scene understanding with action planning. We report the hybrid metrics calculated by the official test server.

\textbf{NuInteract Dataset}~\cite{zhao2025extending} is a large-scale language-based driving dataset containing $\sim$1.5M diverse annotations, including dense captions for individual images and scenes. We utilize NuInteract to establish an initial alignment between BEV visual features and the LLM's semantic space, effectively mitigating data scarcity in vision-language pre-training.

\begin{table}[!t]
\setlength{\tabcolsep}{4.5mm}
\centering
\caption{Training details of \ours. The `/' in Stage 1 and Stage 2 indicates the settings for the two sub-phases.
}
\scriptsize
\label{tab:details}
\begin{tabular}{lccc}
 \toprule
 Config & Stage 1 & Stage 2 & Stage 3\\
 \midrule
 Optimizer & AdamW & AdamW & AdamW\\
 Learning Rate & 2e-4 & 2e-4/4e-4 & 4e-4\\
 Training Epochs & 12/6 & 3/6 & 36\\
 Learning Rate Scheduler & Cosine & Cosine & Cosine\\
 Total Batch Size & 32 & 128 & 128\\
% GPU Device & \multicolumn{3}{c}{32$\times$NVIDIA H20} \\ 
\bottomrule
\end{tabular}
\end{table}

\subsection{Implementation Details}
\label{sec:training-stage}
The BEV-based tokenizer utilizes an OpenCLIP ConvNeXt-L backbone~\cite{liu2022convnet, cherti2023reproducible} for visual feature extraction. Both the visual tokenizer and the Render are initialized from scratch. For the language component, we adopt the pre-trained weights of InternVL2~\cite{chen2024internvl, chen2024far}. The visual tokenizer~\cite{yang2023bevformer} encodes the scene into a $180\times 180$ BEV grid with a channel dimension of 256. In the BEV-to-Point Render, the height dimension $z$ and output channel $c'$ are both set to 32. The model forecasts scene evolution over a horizon of $\Delta t = 3$ seconds. Regarding the training objective, we empirically set the frame-wise generation weights to $\lambda_i = 1 + 0.5 \times i$ for $i \in \{0, \cdots, 3\}$ to emphasize long-term prediction accuracy.

The training process proceeds in three progressive stages, as summarized in Tab.~\ref{tab:details}. First, to establish reliable geometric representations, we conduct a geometry-aware pre-training where the sparse 3D encoder~\cite{yan2018second} is trained via self-supervised point cloud reconstruction and subsequently frozen as a static prior. We then pre-train the tokenizer and Render to lift 2D observations into 3D space by reconstructing current point clouds from multi-view images, supervised by $\mathcal{L}_{\text{render}}$ and $\mathcal{L}_{\text{cos}}$. Subsequently, we bridge the visual and linguistic modalities through alignment and refinement phases while maintaining implicit regularization. We initially train only the LLM projectors using a masking-based augmentation strategy (stitching captions with unmasked views) to mitigate data scarcity, which expands the image-text pairs by $7\times$ to $\sim$200K. This is followed by a refinement phase where all parameters are unfrozen (applying LoRA~\cite{hu2021lora} to the LLM) using dense captions from NuInteract~\cite{zhao2025extending} and scene descriptions from OmniDrive-nuScenes~\cite{wang2024omnidrive}. Finally, we integrate the Current-to-Future Link to conduct scene understanding and predict future evolution. The model is jointly trained on nuScenes keyframes, descriptions, and conversation annotations, guided by Joint Geometric Optimization using both $\mathcal{L}_{\text{gram}}$ and $\mathcal{L}_{\text{cos}}$.

\begin{table*}[t]
\setlength{\tabcolsep}{1.9mm}
\centering
\caption{The comparison of \ours~and understanding/generation specialist models. L/C/T refers to LiDAR/camera/text, respectively. `Aux. Sup.' denotes auxiliary supervision (e.g., 3D object detection, lane detection). We report METEOR, ROUGE, and CIDEr for understanding tasks, and Chamfer Distance for 0--3s on the OmniDrive-nuScenes~\cite{wang2024omnidrive} validation set.}
\scriptsize
\label{tab:main}
\begin{tabular}{ lccccccccccc }
   \toprule
 \multirow{2.3}{*}{Method} & \multirow{2.3}{*}{Reference} & \multirow{2.3}{*}{\# LLM Params}& \multirow{2.3}{*}{Modality}& \multirow{2.3}{*}{Aux. Sup.}&\multicolumn{4}{c}{Generation} & \multicolumn{3}{c}{Understanding}\\
\cmidrule(lr){6-9}\cmidrule(lr){10-12}
 & & & & & 0s $\downarrow$ & 1s $\downarrow$ & 2s $\downarrow$ & 3s $\downarrow$ & METEOR $\uparrow$ & ROUGE $\uparrow$ & CIDEr $\uparrow$ \\
\midrule
\multicolumn{12}{c}{\textit{Only Generation}} \\
\midrule
SPFNet~\cite{weng2021inverting}& CoRL 21 & - & L$\to$L& - &-&2.24 &- &2.50&\multicolumn{3}{c}{\multirow{5}{*}{Unsupported}}\\
S2Net~\cite{weng2022s2net}& ECCV 22 & - & L$\to$L& - &-&1.70 &- &2.06&&&\\
4D-Occ~\cite{khurana2023point}& CVPR 23 & -&L$\to$L& - &-&1.13 &1.53 &2.11&&&\\
ViDAR~\cite{yang2024visual}& CVPR 24 & - &C$\to$L& -&- & 1.12 & 1.38& 1.73&&&\\
DriveX~\cite{shi2025drivex}& ICCV 25 & - & C$\to$L& -&- & \textbf{0.66} & 0.86& 1.10&&&\\
\midrule
\multicolumn{12}{c}{\textit{Only Understanding}} \\
\midrule
LLaVA-OV~\cite{li2024llavaov}& TMLR 25 & 7B&C$\to$T& -&\multicolumn{4}{c}{\multirow{6}{*}{Unsupported}}& -& 0.221 & 0.284\\
Omni-L~\cite{wang2024omnidrive} & CVPR 25 & 7B&C$\to$T& 3D Box, Lane&&&&& 0.376 & 0.321 & 0.732\\
Omni-Q~\cite{wang2024omnidrive} & CVPR 25 & 7B&C$\to$T& 3D Box, Lane&&&&& 0.380 & 0.326 & 0.686\\
OmniDrive-2D~\cite{wang2024omnidrive} & CVPR 25 & 7B&C$\to$T& 3D Box, Lane&&&&& 0.383 & 0.325 & 0.671\\
OmniDrive-BEV~\cite{wang2024omnidrive} & CVPR 25 & 7B&C$\to$T& 3D Box, Lane&&&&& 0.356 & 0.278 & 0.595\\
ORION~\cite{fu2025orion}& ICCV 25 & 7B & C$\to$T& 3D Box, Lane&&&&& 0.354 & 0.306 & 0.635\\
\midrule
\multicolumn{12}{c}{\textit{Unified Understanding and Generation}} \\
\midrule
\oursiccv~\cite{zhou2025hermes} (conference version) & ICCV 25 & 1.8B &C$\to$T\&L& -& 0.59 & 0.78 & 0.95 & 1.17 & 0.384 & 0.327 & 0.741\\
\ours~(\textbf{ours}) & - & 1.8B &C$\to$T\&L& -& 0.53 & 0.71 & 0.86 & 1.01 & 0.385 & 0.327 & 0.749\\
\ours~(\textbf{ours}) & - & 3.8B &C$\to$T\&L& -& \textbf{0.51} & 0.68 & \textbf{0.82} & \textbf{0.97} & \textbf{0.389} & \textbf{0.331} & \textbf{0.772}\\
\bottomrule
\end{tabular}
\end{table*}

\section{Results and Analysis}
In this section, we conduct comprehensive experiments to validate the effectiveness of \ours.

\subsection{Unification of Understanding and Generation}

In this subsection, we present a comprehensive evaluation of \ours~on the NuScenes and OmniDrive-nuScenes benchmarks. We compare our framework against three categories of baselines: 1) Generation specialists; 2) Understanding specialists; and 3) Our conference version. Tab.~\ref{tab:main} details the quantitative comparisons, from which we draw the following observations:

\textbf{1) \ours~significantly advances 3s future scene generation compared to dedicated specialists.}
The quantitative results in Tab.~\ref{tab:main} demonstrate that our method achieves superior geometric prediction accuracy. Compared to leading generation methods such as 4D-Occ~\cite{khurana2023point} and ViDAR~\cite{yang2024visual}, \ours~achieves better performance with only current-frame observations. Specifically, we reduce the CD at the 3s horizon by 41.6\% compared to ViDAR. Even against the recently proposed DriveX~\cite{shi2025drivex}, our method maintains an advantage, achieving a reduction of 0.09 in CD at the 3s horizon. By employing multiple groups of world queries to inject conditional context for future BEV states, our model effectively leverages semantic information to achieve more precise scene evolution prediction.

\textbf{2) \ours~achieves highly competitive understanding capabilities without auxiliary supervision.}
In the domain of 3D scene understanding, \ours~consistently outperforms specialist models while maintaining high data efficiency. As demonstrated in OmniDrive~\cite{wang2024omnidrive}, incorporating auxiliary supervision (e.g., 3D object detection and lane detection) enhances the model's semantic capabilities. However, as indicated in Tab.~\ref{tab:main}, \ours~attains superior performance solely through the BEV representation and standard instruction tuning, without incorporating any detection or map-based supervision. Specifically, we outperform Omni-L and OmniDrive-2D by 2.3\% and 11.6\% on the CIDEr metric, respectively, with consistent gains in METEOR and ROUGE scores. We attribute this improvement to the geometric properties of the BEV representation and the proposed task interaction mechanisms. Baselines such as Omni-Q and ORION typically utilize sparse queries (e.g., via Q-Former3D~\cite{li2023blip}) to extract scene information. While effective, these approaches often benefit from auxiliary supervision to guide feature learning, compensating for the limited geometric context captured by sparse observations. In contrast, \ours~utilizes BEV, which inherently preserves rich semantic information and geometric interaction, allowing the model to learn effective scene representations without external guidance.

Furthermore, we systematically compared \ours with our conference version~\cite{zhou2025hermes}. While the conference version already achieves highly competitive results, \ours~establishes a new state-of-the-art across both tasks. As shown in Tab.~\ref{tab:main}, these upgrades yield a 13.7\% reduction in 3s generation error alongside improvements in understanding metrics, demonstrating the effectiveness of the deeper task interaction enabled by the newly introduced technical improvements. 

Moreover, our framework benefits from model scaling. Increasing the LLM parameters to 3.8B yields improvements across both domains, further reducing the generation error to 0.97 and boosting the CIDEr score to 0.772. This positive correlation indicates that our unified modeling approach effectively internalizes spatial-semantic representations to interpret complex driving scenes, thereby bridging the gap between geometric perception and linguistic understanding.

\begin{figure*}[!th]
	\begin{center}
	\includegraphics[width=0.95\linewidth]{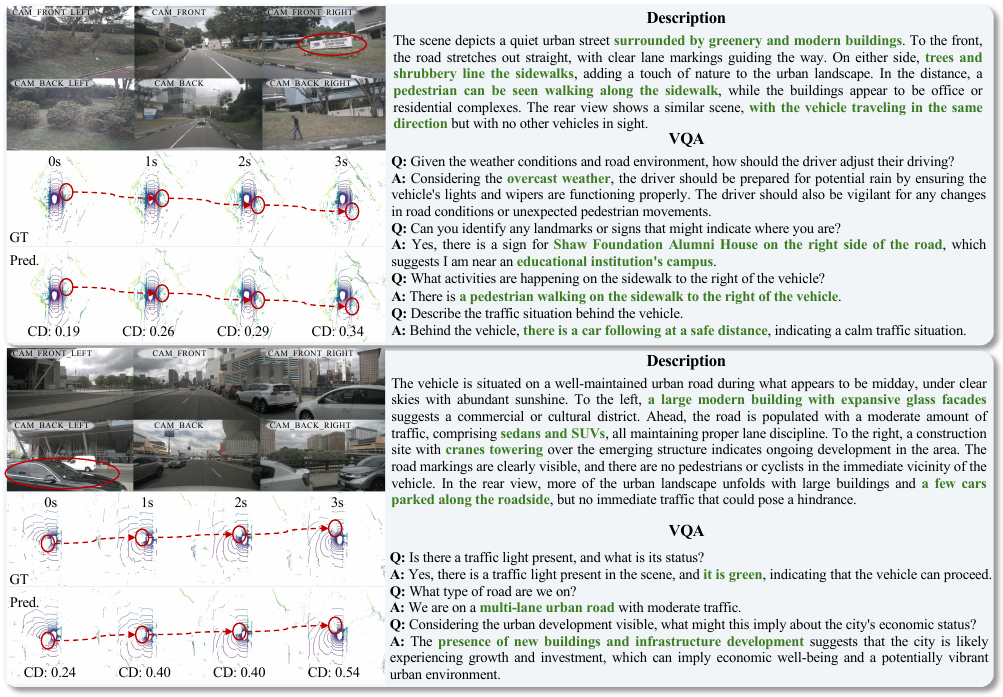}
	\end{center}
	\caption{
   Qualitative results of \ours. The \textcolor[RGB]{59,125,35}{\textbf{green text}} highlights the accurate responses to user instructions. The \textcolor[RGB]{192,0,0}{\textbf{red circles}} track the geometric evolution of other objects in the predicted point clouds.
}
\label{fig:visualization}
\end{figure*}

\subsection{Qualitative Evaluation}
Fig.~\ref{fig:visualization} presents qualitative results from \ours. As highlighted by the green text, the model demonstrates strong scene understanding capabilities. Notably, it is able to identify fine-grained cues such as the ``Shaw Foundation Alumni House'' on signage and infer that the scene is likely located on a campus. This suggests that our BEV representation preserves sufficient granularity for fine-grained semantic reasoning, overcoming the resolution limitations often attributed to compressed features. Regarding scene evolution, the red circles track the precise geometric progression of other objects, maintaining high structural integrity consistent with the ground truth. This alignment between textual interpretation and physical trajectories validates the effectiveness of our framework in bridging semantic understanding with geometric simulation.

\subsection{Analysis and Ablation Study}
The following experiments are conducted on the NuScenes dataset. Unless specified, all the ablation experiments are performed using 25\% of the training data.

\subsubsection{\textbf{Analysis of BEV Input Representation}}
\label{sec:ablation_bev}

We first investigate the BEV input representation, focusing on its superiority over direct multi-view inputs and on how spatial resolution balances geometry information with computational efficiency.

\textbf{Effectiveness of the BEV Input.} To demonstrate the effectiveness of BEV representations as a unified interface, we compare our approach against a strong multi-view baseline where CLIP-encoded image features are directly input into the LLM. To ensure comparable information capacity, image features are resized to 2,532 tokens, a token budget comparable to that of our BEV input. The LLM, BEVFormer, and render head (without up/downsampling layers) then process these features for scene understanding and future prediction.

\begin{figure*}[t]
    \centering
    \includegraphics[width=0.98\linewidth]{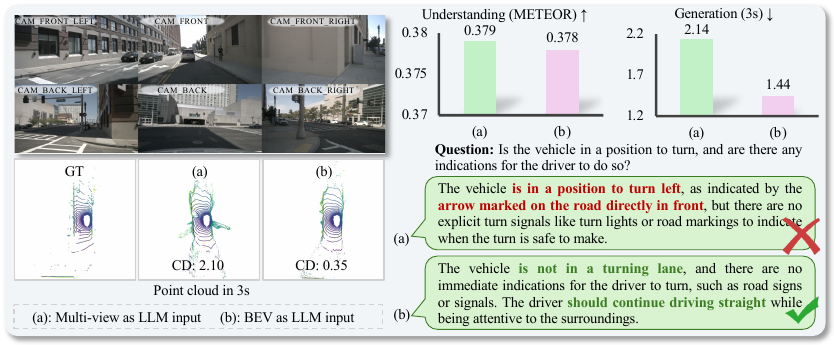}
    \caption{
    Qualitative case and comparison between Multi-view-based and BEV-based inputs. While both methods yield comparable scene understanding performance (see METEOR scores), the multi-view baseline suffers from spatial structural collapse.
    }
    \label{fig:bev_compare}
\end{figure*}

Quantitative results in Fig.~\ref{fig:bev_compare} reveal that while both methods achieve highly competitive scene understanding performance with only 0.001 METEOR difference, they diverge significantly in generation tasks. Our BEV approach significantly outperforms the multi-view baseline, reducing the Chamfer Distance at 3s by $\sim$32\%. We attribute this to the spatial structural collapse of flattened image tokens during LLM processing, which hinders accurate 3D geometry recovery while retaining rich semantics. As shown in the qualitative case in Fig.~\ref{fig:bev_compare}, the multi-view baseline hallucinates a left turn due to misleading road markings, whereas our BEV representation, preserving spatial topology, correctly predicts a straight trajectory. This highlights BEV features as a superior, unified representation that effectively balances semantics with the geometric structural prior required for accurate forecasting.

\begin{table}[t]
    \centering
    \caption{Latency of the vision path before the LLM and the resulting token cost for the two input interfaces. Vision path total covers the image backbone and the feature-to-token stage. All values are measured on one A100 under bf16 and averaged over repeated samples.}
    \label{tab:runtime-bev}
    \scriptsize
    \setlength\tabcolsep{3.6mm}
    \begin{tabular}{lcc}
        \toprule
        Stage & \makecell{Multi-view\\as LLM input} & \makecell{BEV\\as LLM input} \\
        \midrule
        Image backbone, 6 cams $1600{\times}928$ & 112.1\,ms & 112.1\,ms \\
        Feature to LLM tokens & 0.9\,ms & 60.2\,ms \\
        \# LLM visual tokens & 2,532 & 2,025 \\
        LLM prefill & 19.2\,ms & 15.8\,ms \\
        \midrule
        Vision path total & $\sim$113\,ms & $\sim$172\,ms \\
        \bottomrule
    \end{tabular}
\end{table}

\textbf{Inference Cost of the BEV Input.} We further profile the inference cost of the two input interfaces. Both variants are measured on a single A100 under the same setting, with bf16 autocast enabled in the vision path. Since they share the same image backbone and are given comparable token budgets, the BEVFormer and the downsampling layer account for the entire runtime difference. Tab.~\ref{tab:runtime-bev} shows that the attention over BEV grids is not a dominant cost: the BEVFormer and the downsampling layer add about 60\,ms, which brings the vision path before the LLM from about 113\,ms to about 172\,ms, while the token budget and the prefill of the two interfaces remain comparable. This cost is incurred once per frame and is independent of the length of the generated answer and of the number of predicted future frames. We regard it as an acceptable price for an interface that accommodates spatial geometry and semantics at the same time, which is what yields the substantial gain in geometric accuracy reported above.

\textbf{Impact of BEV Representation Scales.} 
We further evaluate the impact of BEV spatial resolution, which entails a trade-off between geometric consistency and the token-processing capacity of LLMs. As shown in Tab.~\ref{tab:ablation-bevsize}, we compare downsampling strategies ($\times 4, \times 8$) against directly learning coarse features (`Direct Query'). The `Downsample ($\times 4$)' yields the best trade-off between generation and understanding, achieving the lowest CD at 0--3s and a CIDEr score of 0.720. Notably, downsampling from a fine-grained feature map significantly surpasses direct coarse querying. Specifically, the `Downsample ($\times 4$)' strategy reduces the CD at 3s from 2.012 to 1.436, representing a 26.8\% improvement. This result indicates that capturing detailed geometric information at a high resolution prior to compression is essential for preserving structural integrity, whereas direct coarse querying results in irretrievable information loss. In contrast, excessive downsampling ($\times 8$) creates an information bottleneck, degrading the CD at 3s to 1.781 and dropping the CIDEr to 0.681, confirming that sufficient spatial resolution is critical for both precise future prediction and semantic understanding.

\begin{table}[t]
    \centering
    \caption{Ablation on downsampling scales for the BEV. M., R., and C. are METEOR, ROUGE, and CIDEr, respectively.}
    \label{tab:ablation-bevsize}
    \scriptsize
    \setlength\tabcolsep{1.8mm}
    \begin{tabular}{lccccccccc}
        \toprule
        \multirow{2.3}{*}{BEV Configuration} & \multicolumn{4}{c}{Generation} & \multicolumn{3}{c}{Understanding}\\
        \cmidrule(lr){2-5}\cmidrule(lr){6-8}
        & 0s $\downarrow$ & 1s $\downarrow$ & 2s $\downarrow$ & 3s $\downarrow$ & M. $\uparrow$ & R. $\uparrow$ & C. $\uparrow$ \\
        \midrule
        Downsample ($\times 8$) & 0.827 & 1.084 & 1.410 & 1.781 & 0.370 & 0.314 & 0.681 \\
        Downsample ($\times 4$) & 0.588 & 0.879 & 1.146 & 1.436 & 0.378 & 0.322 & 0.720 \\
        Direct Query ($\times 4$) & 0.769 & 1.211 & 1.615 & 2.012 & 0.378 & 0.322 & 0.723 \\
        \bottomrule
    \end{tabular}
    % \vspace{-5pt}
\end{table}

\subsubsection{\textbf{Analysis of Joint Geometric Optimization}}

\begin{table}[!t]
\setlength{\tabcolsep}{2.2mm}
\centering
\caption{Ablation on the Joint Geometric Optimization strategy. M., R., and C. indicate METEOR, ROUGE, and CIDEr, respectively.}
\scriptsize
\label{tab:ablation-align}
\begin{tabular}{ccccccccc}
    \toprule
    \multirow{2.3}{*}{$\mathcal{L}_{\text{cos}}$} & \multirow{2.3}{*}{$\mathcal{L}_{\text{gram}}$} & \multicolumn{4}{c}{Generation} & \multicolumn{3}{c}{Understanding}\\
    \cmidrule(lr){3-6}\cmidrule(lr){7-9}
    && 0s $\downarrow$ & 1s $\downarrow$ & 2s $\downarrow$ & 3s $\downarrow$ & M. $\uparrow$ & R. $\uparrow$ & C. $\uparrow$ \\
    \midrule
        -      & - & 0.656 & 0.967 & 1.270 & 1.637 & 0.379 & 0.322 & 0.722 \\ 
    \checkmark & - & 0.594 & 0.890 & 1.161 & 1.441 & 0.378 & 0.321 & 0.717 \\
    - &\checkmark  & 0.608 & 0.920 & 1.218 & 1.544 & 0.378 & 0.321 & 0.717 \\
     \checkmark &\checkmark & 0.588 & 0.879 & 1.146 & 1.436 & 0.378 & 0.322 & 0.720 \\
    \bottomrule
\end{tabular}
\end{table}

\begin{figure}[t]
	\begin{center}
	\includegraphics[width=0.99\linewidth]{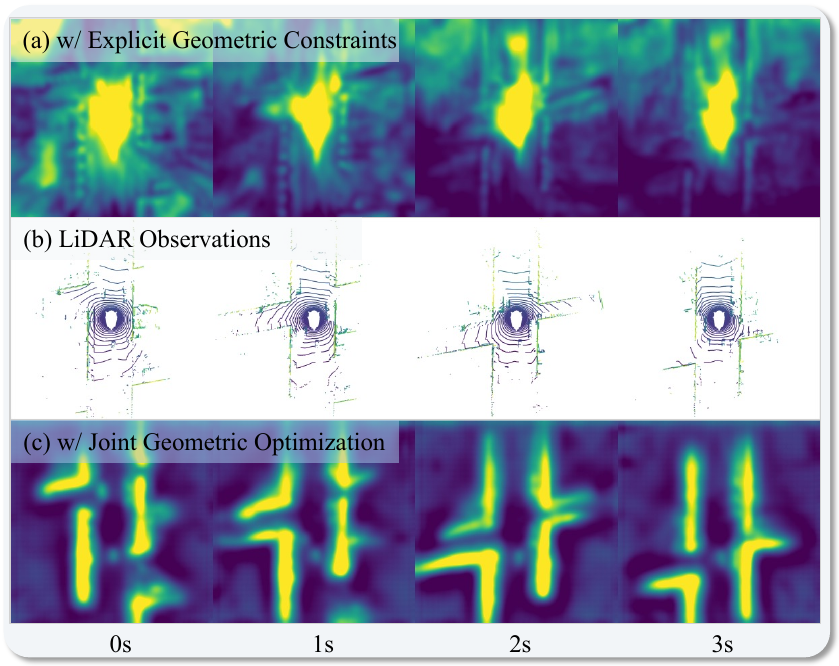}
	\end{center}
	\caption{
   Visualization of internal representations. (a) Features learned with only explicit geometric constraints. (b) The corresponding ground-truth point cloud. (c) Predicted features learned with the Joint Geometric Optimization strategy.
}
\label{fig:viz_feat}
\end{figure}

We then investigate the impact of the proposed Joint Geometric Optimization strategy. As in Tab.~\ref{tab:ablation-align}, depending exclusively on explicit geometric constraints yields a suboptimal CD of 1.637 at the 3s horizon. Incorporating implicit geometric regularization with $\mathcal{L}_{\text{cos}}$ significantly enhances performance, reducing the CD to 1.441. This indicates that enforcing voxel-wise feature consistency with geometry-aware priors is essential for accurate structure generation. Furthermore, only employing $\mathcal{L}_{\text{gram}}$ also provides a noticeable improvement (CD of 1.544). We attribute this to the ability of the Gram matrix to capture global structural patterns and internal feature correlations, serving as a complement to the spatially strict Cosine loss. The integration of both implicit geometric regularizers yields superior performance (CD of 1.436), demonstrating a synergistic effect in which joint constraints on local feature consistency and global structural coherence encourage plausible 3D representations while maintaining understanding performance.

To further validate the geometric regularization effect, we visualize the learned BEV features and demonstrate the effectiveness of our Joint Geometric Optimization strategy. As shown in Fig.~\ref{fig:viz_feat}(a), relying on explicit geometric constraints inevitably leads to depth ambiguity. This manifests as conspicuous ray-shaped artifacts extending along camera projection lines and an abnormally high response concentration at the ego-center, which overshadows the scene's essential geometric structure. In contrast, the proposed Joint Geometric Optimization strategy imposes explicit constraints on the output while implicitly regularizing the latent manifold. As shown in Fig.~\ref{fig:viz_feat}(c), this approach effectively suppresses projection artifacts and the central bias, yielding spatially compact features that strictly adhere to the intrinsic geometry presented in the point cloud (Fig.~\ref{fig:viz_feat}(b)). This confirms that the joint mechanism guides the model to learn a cleaner, structurally faithful latent space rather than overfitting to perspective camera views.

\subsubsection{\textbf{Analysis of Current-to-Future Link}}
We perform a progressive ablation study to validate our Current-to-Future Link in Tab.~\ref{tab:ablation-link}. The `w/o Link' entirely removes the Link module, attempting to predict the future by directly copying $\mathbf{B}_t$ and adding future ego-motion states. This naive propagation fails to model the dynamic evolution of the environment, resulting in a severe performance drop with a 3s CD of 2.377 and a CIDEr drop to 0.433. Introducing the Simple Link, consisting of 3 vanilla attention layers, resolves this bottleneck and drastically reduces the 3s CD to 1.542.

Incorporating Textual Injection reduces the CD to 1.506, suggesting that semantic abstractions and world knowledge derived from the understanding branch provide crucial context for geometric generation, effectively conditioning future prediction on linguistic priors. Another pronounced improvement is achieved by introducing the Ego Modulation, which further lowers the CD to 1.442. This module injects the ego-vehicle's kinematic states into the feature modulation process. It is pivotal for decoupling the ego-motion from the scene's inherent dynamics, thereby preventing the misinterpretation of static background shifts as object motion. Finally, scaling up the network depth from 3 to 6 achieves the best CD of 1.436 and boosts the CIDEr score to 0.720. This indicates that sufficient model capacity is required to capture the highly non-linear temporal evolution of complex driving scenarios. At the same time, the parallel improvement in understanding metrics suggests a positive interaction between the two tasks. Compared with the implicit geometric regularization, which reduces the 3s CD from 1.637 to 1.436 (Tab.~\ref{tab:ablation-align}), the gains of the Textual Injection and the Ego Modulation are smaller but consistent across horizons.

\subsubsection{\textbf{Analysis of task interaction}}
\begin{table}[!t]
\setlength{\tabcolsep}{1.9mm}
\centering
\caption{Ablation on Current-to-Future Link. M., R., and C. indicate METEOR, ROUGE, and CIDEr, respectively.}
\scriptsize
\label{tab:ablation-link}
\begin{tabular}{lccccccc}
    \toprule
    \multirow{2.3}{*}{Modules} & \multicolumn{4}{c}{Generation} & \multicolumn{3}{c}{Understanding}\\
    \cmidrule(lr){2-5}\cmidrule(lr){6-8}
    & 0s $\downarrow$ & 1s $\downarrow$ & 2s $\downarrow$ & 3s $\downarrow$ & M. $\uparrow$ & R. $\uparrow$ & C. $\uparrow$ \\
    \midrule
     w/o Link & 0.891 & 1.368 & 1.877 & 2.377 & 0.290 & 0.263 & 0.433 \\
     w/ Simple Link & 0.609 & 0.919 & 1.207 & 1.542 & 0.378 & 0.322 & 0.718 \\
     + Textual Injection & 0.602 & 0.907 & 1.205 & 1.506 & 0.376 & 0.321 & 0.717 \\ 
     + Ego Modulation & 0.597 & 0.895 & 1.165 & 1.442 & 0.377 & 0.320 & 0.711 \\
     + More blocks  & \textbf{0.588} & \textbf{0.879} & \textbf{1.146} & \textbf{1.436} & \textbf{0.378} & \textbf{0.322} & \textbf{0.720} \\
    \bottomrule
\end{tabular}
\end{table}

\begin{table}[t]
    \centering
    \caption{Ablation on interaction of tasks. M., R., and C. indicate METEOR, ROUGE, and CIDEr, respectively.}
    \scriptsize
    \label{tab:ablation-relation}
    \setlength\tabcolsep{1.8mm}
    \begin{tabular}{ccccccccccc}
        \toprule
        \multirow{2.3}{*}{Under.} & \multirow{2.3}{*}{Gen.} & \multicolumn{4}{c}{Generation} & \multicolumn{3}{c}{Understanding}\\
        \cmidrule(lr){3-6}\cmidrule(lr){7-9}
        & & 0s $\downarrow$ & 1s $\downarrow$ & 2s $\downarrow$ & 3s $\downarrow$ & M. $\uparrow$ & R. $\uparrow$ & C. $\uparrow$ \\
        \midrule
        \textcolor{gray}{\checkmark} & \textcolor{gray}{-} & \textcolor{gray}{-} & \textcolor{gray}{-} & \textcolor{gray}{-} & \textcolor{gray}{-} & \textcolor{gray}{0.379} & \textcolor{gray}{0.322} & \textcolor{gray}{0.726} \\
        \textcolor{gray}{-}& \textcolor{gray}{\checkmark}& \textcolor{gray}{0.599} & \textcolor{gray}{0.885} & \textcolor{gray}{1.157} & \textcolor{gray}{1.434} &\textcolor{gray}{-}&\textcolor{gray}{-}&\textcolor{gray}{-} \\
        \midrule
        \multicolumn{2}{c}{Separated unification} & 0.600 & 0.981 & 1.296 & 1.634 & 0.376 & 0.320 & 0.703 \\
        \checkmark & \checkmark & \textbf{0.588} & \textbf{0.879} & \textbf{1.146} & \textbf{1.436} & \textbf{0.378} & \textbf{0.322} & \textbf{0.720} \\
        \bottomrule
    \end{tabular}
\end{table}

We then analyze the interaction between future generation and scene understanding to validate the necessity of the proposed framework.

\textbf{Impact of Task Unification.} 
Tab.~\ref{tab:ablation-relation} evaluates the influence of joint training compared to single-task specialists. The `Separated unification' baseline, where tasks share the visual tokenizer but operate without deep interaction, yields suboptimal performance with a CD of 1.634 at 3 seconds and a CIDEr score of 0.703. In contrast, our framework significantly improves these metrics to 1.436 and 0.720, respectively. This demonstrates that the proposed unified modeling is effective. Specifically, the semantic context guides geometric evolution, while geometric constraints provide physical grounding for language reasoning. Thus, the joint model achieves performance comparable to specialized single-task models without requiring separate architectures.

\textbf{Impact of World Queries Integration Strategy.} 
We further analyze the integration of world queries $\mathbf{Q}^w$ with the LLM. We compare three settings as shown in Tab.~\ref{tab:ablation-worldquery}. The baseline (a), i.e., separated unification, which relies solely on Textual Injection for task unification, yields a suboptimal CD of 1.634. Setting (b) introduces $\mathbf{Q}^w$ to aggregate BEV features but bypasses the LLM reasoning process. While this improves the CD to 1.526 by enhancing feature extraction, the lack of deep semantic interaction limits its potential. Our proposed method (c), which processes $\mathbf{Q}^w$ directly through the LLM, achieves the best performance with a CD of 1.436 and a CIDEr score of 0.720. This improvement indicates that the LLM effectively encodes semantic information into the $\mathbf{Q}^w$, which serves as a critical guide for future geometric prediction. Furthermore, since the world queries participate in the forward pass, the gradients from the generation objective backpropagate to refine the LLM, thereby improving the text understanding capability.

\begin{table}[t]
    \centering
    \caption{Ablation on world queries integration. M., R., and C. indicate METEOR, ROUGE, and CIDEr, respectively.}
    \scriptsize
    \label{tab:ablation-worldquery}
    \includegraphics[width=\linewidth]{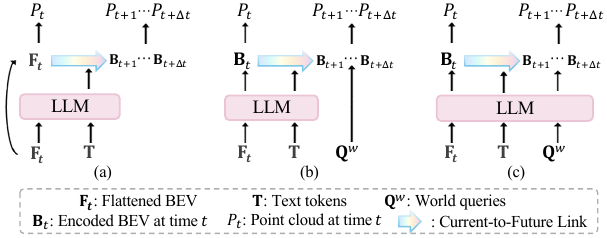}
    
    \vspace{2mm}

    \setlength\tabcolsep{2.5mm}
    \begin{tabular}{lccccccc}
        \toprule
        \multirow{2.3}{*}{Setting} & \multicolumn{4}{c}{Generation} & \multicolumn{3}{c}{Understanding}\\
        \cmidrule(lr){2-5}\cmidrule(lr){6-8}
        & 0s $\downarrow$ & 1s $\downarrow$ & 2s $\downarrow$ & 3s $\downarrow$ & M. $\uparrow$ & R. $\uparrow$ & C. $\uparrow$ \\
        \midrule
        (a)& 0.600 & 0.981 & 1.296 & 1.634 & 0.376 & 0.320 & 0.703 \\
        (b)& 0.597 & 0.920 & 1.202 & 1.526 & 0.377 & 0.320 & 0.709 \\
        (c) & \textbf{0.588} & \textbf{0.879} & \textbf{1.146} & \textbf{1.436} & \textbf{0.378} & \textbf{0.322} & \textbf{0.720} \\
        \bottomrule
    \end{tabular}
\end{table}

\begin{table}[t]
    \centering
    \caption{Ablation on hyperparameters and configurations. M., R., and C. indicate METEOR, ROUGE, and CIDEr.}
    \label{tab:ablation_all}
    \scriptsize    
    
    \begin{subtable}[t]{\linewidth}
        \setlength{\tabcolsep}{2.2mm} 
        \centering
        \caption{Ablation on the source of world queries}
        \label{tab:ablation-init}
        \begin{tabular}{lccccccc}
        \toprule
        \multirow{2.3}{*}{Method} & \multicolumn{4}{c}{Generation} & \multicolumn{3}{c}{Understanding}\\
        \cmidrule(lr){2-5}\cmidrule(lr){6-8}
        & 0s $\downarrow$ & 1s $\downarrow$ & 2s $\downarrow$ & 3s $\downarrow$ & M. $\uparrow$ & R. $\uparrow$ & C. $\uparrow$ \\
        \midrule
        Random Init& 0.598 & 0.883 & 1.161 & 1.448 & 0.378 & 0.321 & 0.714 \\
        Attn. Pool& 0.595 & 0.892 & 1.152 & 1.438 & 0.377 & 0.321 & 0.716 \\
        Cross Attn.& 0.603 & 0.892 & 1.155 & 1.442 & 0.378 & 0.322 & 0.719 \\
        Avg. Pool& 0.597 & 0.883 & 1.153 & 1.444 & 0.378 & 0.321 & 0.713 \\
        Max Pool & 0.588 & 0.879 & 1.146 & 1.436 & 0.378 & 0.322 & 0.720 \\
        \bottomrule
        \end{tabular}
    \end{subtable}
    
    \vspace{1.5mm}

    \begin{subtable}[t]{\linewidth}
        \setlength{\tabcolsep}{2.9mm} 
        \centering
        \caption{Ablation on the number of world queries}
        \label{tab:ablation-n}
        \begin{tabular}{lccccccc}
            \toprule
            \multirow{2.3}{*}{$n$} & \multicolumn{4}{c}{Generation} & \multicolumn{3}{c}{Understanding}\\
            \cmidrule(lr){2-5}\cmidrule(lr){6-8}
            & 0s $\downarrow$ & 1s $\downarrow$ & 2s $\downarrow$ & 3s $\downarrow$ & M. $\uparrow$ & R. $\uparrow$ & C. $\uparrow$ \\
            \midrule
            0  & 0.598 & 0.910 & 1.173 & 1.478 & 0.377 & 0.321 & 0.716 \\
            1  & 0.593 & 0.888 & 1.146 & 1.419 & 0.377 & 0.321 & 0.712 \\
            2  & 0.599 & 0.884 & 1.148 & 1.431 & 0.377 & 0.321 & 0.717 \\
            4  & 0.588 & 0.879 & 1.146 & 1.436 & 0.378 & 0.322 & 0.720 \\
            8  & 0.594 & 0.889 & 1.152 & 1.430 & 0.378 & 0.321 & 0.719\\
            \bottomrule
        \end{tabular}
    \end{subtable}
    
    \vspace{1.5mm}

    \begin{subtable}[t]{\linewidth}
    \setlength{\tabcolsep}{1.4mm}
    \centering
    \caption{Ablation on generation length. The last row extends the supervision up to 10s. The 10s column is averaged over validation scenarios with ground truth at that horizon.}
    \label{tab:ablation-length}
    \begin{tabular}{cccccccccccc}
        \toprule
        \multicolumn{4}{c}{Second} &\multicolumn{5}{c}{Generation} & \multicolumn{3}{c}{Understanding}\\
        \cmidrule(lr){1-4}\cmidrule(lr){5-9}\cmidrule(lr){10-12}
        0&1&2 & 3 & 0s $\downarrow$ & 1s $\downarrow$ & 2s $\downarrow$ & 3s $\downarrow$ & 10s $\downarrow$ & M. $\uparrow$ & R. $\uparrow$ & C. $\uparrow$ \\
        \midrule
        \checkmark &\checkmark & -& - & 0.550 & 0.835 & - & - & - & 0.380 & 0.322 & 0.723 \\
        \checkmark &\checkmark & \checkmark& - & 0.575 & 0.851 & 1.147 & - & - & 0.378 & 0.322 & 0.718 \\
        - &\checkmark & \checkmark &\checkmark & - & 0.933 & 1.189 & 1.476 & - & 0.378 & 0.321 & 0.716 \\
        \checkmark & - & - &\checkmark & 0.584 & - & - & 1.677 & - & 0.379 & 0.322 & 0.723 \\
        \checkmark &\checkmark & \checkmark &\checkmark & 0.588 & 0.879 & 1.146 & 1.436 & - & 0.378 & 0.322 & 0.720 \\
        \multicolumn{4}{c}{Extended to 10s} & 0.704 & 1.608 & 2.476 & 3.139 & 4.579 & 0.376 & 0.323 & 0.718 \\
        \bottomrule
    \end{tabular}
    \end{subtable}
\end{table}

\subsubsection{\textbf{Analysis of Hyperparameters and Configurations}}

Finally, we systematically investigate the impact of specific network hyperparameters and configurations, including the query initialization strategy, the number of world queries, and the temporal prediction horizon.

\textbf{Impact of Initialization of World Queries.} 
We first investigate the strategy for initializing world queries to validate the necessity of perceptual anchoring. As shown in Tab.~\ref{tab:ablation-init}, we first introduce a `Random Init' baseline where queries are learned as independent parameters. This setting yields a suboptimal CD of 1.448 at 3s. In contrast, initializing queries directly from BEV features consistently improves performance. This confirms that anchoring world queries to the current perceptual state establishes a direct channel between input features and future predictions, facilitating effective gradient flow and semantic alignment. Among the BEV-based strategies, we compare parametric methods (e.g., Attention Pooling, Cross Attention) with heuristic ones (e.g., Average Pooling, Max Pooling). Notably, the simple Max Pooling strategy achieves the best overall performance (CD 1.436), surpassing the more complex Cross Attention mechanism. We attribute this to the inherent sparsity of BEV representations, where Max Pooling effectively captures the most salient features within the grid  (e.g., object occupancy) while ignoring background noise. Average Pooling dilutes these signals, and parametric methods may face optimization challenges due to limited training data.

\textbf{Impact of Number of World Queries.} 
We further analyze the quantity $n$ of world queries $\mathbf{Q}^w \in \mathbb{R}^{(\Delta t \times n) \times C}$, which serve as an information bridge between temporal BEV features and the LLM. As shown in Tab.~\ref{tab:ablation-n}, $n=0$ relies solely on Textual Injection within the Current-to-Future Link, which yields a suboptimal CD of 1.478 at the 3s horizon. Introducing world queries establishes a dedicated geometric-semantic bridge to guide future generation. Specifically, setting $n=4$ reduces the 3s CD by 0.04 while elevating the CIDEr metric to 0.720. This validates the effectiveness of the queries in aggregating semantic context from BEV observations and textual instructions. Further increasing $n$ to 8 causes slight regressions in 0--2s generation. Thus, we adopt $n$=4 as the default configuration to balance computational efficiency with representational capacity.

\textbf{Impact of Generation Horizon.} 
Tab.~\ref{tab:ablation-length} presents the impact of prediction horizon length on model performance. While shorter horizons such as 0--1s yield a lower CD of 0.550 due to reduced uncertainty, modeling long-term evolution is essential for comprehensive scene understanding. We observe that temporal continuity is critical. For instance, the discontinuous setting of 0s and 3s results in a sharp degradation in CD from 3s to 1.677, indicating that intermediate states serve as necessary bridges for reasoning about future dynamics. Similarly, excluding the current state (1--3s) hampers performance as the model lacks the initial geometric reference. Our default setting of 0--3s strikes an optimal balance, maintaining long-term prediction with a CD of 1.436 while preserving high semantic understanding capabilities.

As shown in the last row of Tab.~\ref{tab:ablation-length}, extending the supervised horizon to 10 seconds maintains understanding metrics but degrades predicted geometry. The CD at 3s increases from 1.436 to 3.139, reaching 4.579 at 10s. This degradation primarily affects the predicted future rather than the current frame, suggesting long-horizon targets are weakly determined. We attribute this behavior to the weak determinacy of long-horizon targets. Within 10s, the ego vehicle often travels beyond the evaluation range, making the geometry largely unobservable from the current frame. The inconsistent gradients from long-horizon supervision interfere with near-horizon objectives, causing a slight degradation in 1-3 second accuracy despite unchanged supervision. This indicates that a metric-based protocol is most reliable within the observable range, justifying our 3-second horizon for main experiments. Reliable long-horizon geometry prediction requires a larger observable context or an evaluation protocol accounting for unobserved regions, which warrants future research.

\subsection{Generalization to Additional Tasks}
\subsubsection{\textbf{Understanding Capability on Other Datasets}}

\begin{table}[t]
\centering
\setlength{\tabcolsep}{3.7mm}
\caption{VQA results on NuScenes-QA dataset~\cite{qian2024nuscenes}.}
\scriptsize
\vspace{-2pt}
\label{tab:nusc_qa}
\begin{tabular}{lcccccccccc}
\toprule
    Method & Reference & Modality & Acc. (\%) $\uparrow$ \\
    \midrule
     LLaVA~\cite{liu2023visual} & NeurIPS 23 & Camera & 47.4 \\
     BEVDet+BUTD~\cite{qian2024nuscenes} & AAAI 24 & Camera & 57.0 \\
     BEVDet+MCAN~\cite{qian2024nuscenes} & AAAI 24 & Camera & 57.9 \\
     CenterPoint+BUTD~\cite{qian2024nuscenes} & AAAI 24 & LiDAR & 58.1 \\
     CenterPoint+MCAN~\cite{qian2024nuscenes} & AAAI 24 & LiDAR & 59.5 \\
     LiDAR-LLM~\cite{yang2025lidar} & AAAI 25 & LiDAR & 48.6 \\
     Omni-Q~\cite{wang2024omnidrive} & CVPR 25 & Camera & 59.2 \\ 
     OpenDriveVLA-7B~\cite{zhou2025opendrivevla} & AAAI 26 & Camera & 58.2 \\
\midrule
\ours & - & Camera & \textbf{61.3} \\
\bottomrule
\end{tabular}
\end{table}

\begin{table}[t]
    \centering
    \scriptsize
    \caption{The results on driving with language leaderboard~\cite{sima2024drivelm}, where `Acc.', `GPT', and `FS' indicate Accuracy, GPT-score, and Final Score, respectively.} 
    \setlength\tabcolsep{2.3mm}
    \label{tab:drivelm}
    \begin{tabular}{lccccc}
        \toprule
        Method & Reference & Acc. $\uparrow$ & GPT $\uparrow$ & Match $\uparrow$ & FS $\uparrow$ \\
        \midrule
        DriveLM~\cite{sima2024drivelm} & ECCV 24 & 0.65 & 0.53 & 0.28 & 0.50 \\
        Team NVIDIA~\cite{adchallenge2024} & CVPRW 24 & 0.78 & 0.60 & - & 0.59 \\
        MMFM\_AD~\cite{adchallenge2024} & CVPRW 24 &0.67 & 0.64 & - & 0.57 \\
        Omni-Q~\cite{wang2024omnidrive} & CVPR 25 & 0.78 & \textbf{0.64} & 0.37 & 0.58 \\
        FSDrive~\cite{zeng2025futuresightdrive} & NeurIPS 25 & 0.72 & 0.63 & 0.39 & 0.57 \\
        \midrule
        \ours & - & \textbf{0.83} & 0.61 & \textbf{0.43} &\textbf{0.59} \\
        \bottomrule
    \end{tabular}
\end{table}

To further evaluate the generalization capability of \ours~in diverse driving scenarios, we extend our evaluation to two popular benchmarks: NuScenes-QA~\cite{qian2024nuscenes}, which focuses on 3D spatial perception statistics, and DriveLM~\cite{sima2024drivelm}, which emphasizes graph-based reasoning.

As shown in Tab.~\ref{tab:nusc_qa}, \ours~achieves state-of-the-art performance with an accuracy of 61.3\%. Generalist 2D VLMs such as LLaVA~\cite{liu2023visual} struggle on this benchmark, largely due to their lack of 3D spatial modeling. In contrast, modular approaches that combine 3D detectors with QA heads, such as CenterPoint+MCAN~\cite{qian2024nuscenes}, perform significantly better. Notably, \ours~outperforms the camera-based SOTA Omni-Q~\cite{wang2024omnidrive} by 2.1\% and effectively surpasses the LiDAR-based specialist CenterPoint+MCAN, which scores 59.5\%. This result demonstrates that the BEV representation effectively encodes rich geometric information comparable to LiDAR, providing robust spatial grounding for visual question answering without relying on depth sensors during inference.

We further evaluate the reasoning capability of \ours~on the DriveLM benchmark, which demands integrated reasoning across perception, prediction, and planning. As shown in Tab.~\ref{tab:drivelm}, \ours~achieves a highly competitive Final Score (FS) of 0.59, matching the challenge winner Team NVIDIA and outperforming strong baselines like Omni-Q~\cite{wang2024omnidrive} and FSDrive~\cite{zeng2025futuresightdrive}. Notably, as shown in Fig.~\ref{fig:intro}(d), our method achieves a prediction accuracy of 0.83, surpassing Omni-Q (0.78) by 5\%. Furthermore, \ours~attains a leading score of 0.43 in the `Match' metric without auxiliary detection supervision~\cite{wang2023exploring}, validating the advantage of the 3D representation. By enforcing geometric consistency through the generation branch, the model inherently encodes precise spatial semantics, enabling it to ground its reasoning in a 3D scene rather than relying solely on language priors.

\begin{table}[!t]
\centering
\caption{
Comparison of the motion planning in nuScenes~\cite{caesar2020nuscenes} validation set.}
\footnotesize
\setlength\tabcolsep{.65mm} %列
\begin{tabular}{lccccccccc}
\toprule
\multirow{2.3}{*}{Method} &
\multirow{2.3}{*}{Reference} &
\multicolumn{4}{c}{L2 (m) $\downarrow$} & 
\multicolumn{4}{c}{Collision (\%) $\downarrow$}\\ 
\cmidrule(lr){3-6} \cmidrule(lr){7-10}
 && 1s & 2s & 3s &Avg. & 1s & 2s & 3s& Avg.\\
        \midrule
        Ego-MLP~\cite{zhai2023rethinking} & arXiv 23& 0.15 & 0.32 & 0.59  & 0.35&0.00 & 0.27 & 0.85&0.37\\
        BEV-Planner~\cite{li2024ego} &  CVPR 24& 0.30 & 0.52&0.83 &0.55 & 0.10 & 0.37 & 1.30 &0.59\\
        BEV-Planner++~\cite{li2024ego} & CVPR 24& 0.16 & 0.32& 0.57 & 0.35& 0.00 & 0.29 & 0.73 &0.34 \\
        \midrule
        ST-P3~\cite{hu2022st} &  ECCV 22 & 1.33 & 2.11 & 2.90 & 2.11 & 0.23 & 0.62 & 1.27 & 0.71 \\
        UniAD~\cite{hu2023planning} & CVPR 23 & 0.20 & 0.42 & 0.75& 0.46 & 0.02 & 0.25 & 0.84&0.37  \\
        VAD-Base~\cite{jiang2023vad} & ICCV 23 & 0.17 & 0.34 & 0.60 &0.37 & 0.04 & 0.27 & 0.67 & 0.33  \\
        OmniDrive~\cite{wang2024omnidrive} & CVPR 25 & 0.40 & 0.80 & 1.32 & 0.84 & 0.04 & 0.46 & 2.32 & 0.94 \\
        OmniDrive++~\cite{wang2024omnidrive} & CVPR 25 & 0.14 & 0.29 & 0.55 & 0.33 & 0.00 & 0.13 & 0.78 & 0.30 \\
        Doe-1~\cite{zheng2024doe}& arXiv 24 & 0.50 & 1.18 & 2.11 & 1.26 & 0.04 & 0.37 & 1.19 & 0.53 \\
        Epona~\cite{zhang2025epona}&  ICCV 25 & 0.61 & 1.17 & 1.98 & 1.25 & 0.01 & 0.22 & 0.85 & 0.36 \\
        ORION~\cite{fu2025orion} & ICCV 25 & 0.17 & 0.31& 0.55 & 0.34& 0.05 & 0.25 & 0.80 & 0.37  \\
        \midrule
        \ours & - & 0.16 & 0.33 & 0.62 & 0.37 & 0.00 & 0.16 & 0.72 & 0.29\\
\bottomrule
\end{tabular}
\label{tab:nuscenes-plan}
\end{table}

\subsubsection{\textbf{Motion Planning}}

To further evaluate the generalization capability of our learned unification model, we extend \ours~to the open-loop motion planning task on the nuScenes validation set. Specifically, we append a lightweight MLP head to the world queries $\mathbf{Q}^w$ to hierarchically regress future trajectories, which subsequently serve as conditions for generating future point clouds. It is worth emphasizing that our model is trained solely with text instructions and future geometric supervision. As shown in Tab.~\ref{tab:nuscenes-plan}, \ours~achieves highly competitive performance with an average L2 error of 0.37m and a collision rate of 0.29\%. Compared to the recent leading method ORION, \ours~achieves a 0.08\% lower collision rate while maintaining a comparable L2 error. Furthermore, our method substantially improves over OmniDrive and surpasses OmniDrive++ in average collision rate. These results demonstrate that by optimizing for future scene generation under textual guidance, \ours~effectively internalizes world knowledge and actionable driving dynamics, enabling planning potential even without perception supervision.

\subsubsection{\textbf{Generalization to Different LLMs}}
To verify the universality of~\ours, we conduct experiments across various LLM architectures and scales using 25\% of the training data.

\textbf{Effect of Model Architecture.} We first evaluate the adaptability of our method by integrating three representative LLMs with comparable parameter counts~\cite{chen2024far,yang2025qwen3,meta2024llama32}. As shown in Tab.~\ref{tab:series-comparison}, our framework consistently achieves promising performance across all architectures, demonstrating its generalization capability. InternVL2 yields superior results in both generation and understanding metrics (e.g., achieving the lowest prediction error at 3s). We attribute this to its reliable vision-language alignment, which effectively adapts to our BEV-based tokenization strategy.

\textbf{Impact of Model Scale.} We further investigate the scalability of our approach using the InternVL2 series. Tab.~\ref{tab:scale-comparison} reveals a clear positive correlation between model size and performance. As the number of parameters increases, the model achieves significant gains in both future prediction and scene understanding. Specifically, the 3.8B model outperforms the 0.8B variant by 12.5\% for the 3s prediction error. We argue that our framework effectively leverages the richer world knowledge and stronger reasoning capabilities inherent in larger language models. This indicates that our method is highly scalable and has the potential for further performance gains as foundation models become increasingly powerful.

\begin{table}[t]
    \centering
    \caption{Comparison of different LLMs. M., R., and C. indicate METEOR, ROUGE, and CIDEr, respectively.}
    \label{tab:combined_analysis}
        \begin{subtable}{1.0\linewidth}
        \centering
        \caption{Comparison across different LLM series.}
        \label{tab:series-comparison}
        \scriptsize
        \setlength\tabcolsep{2.mm}
        \begin{tabular}{lccccccc}
            \toprule
            \multirow{2.3}{*}{Model} & \multicolumn{4}{c}{Generation} & \multicolumn{3}{c}{Understanding}\\
            \cmidrule(lr){2-5}\cmidrule(lr){6-8}
            & 0s $\downarrow$ & 1s $\downarrow$ & 2s $\downarrow$ & 3s $\downarrow$ & M. $\uparrow$ & R. $\uparrow$ & C. $\uparrow$ \\
            \midrule
            Llama-3.2~\cite{meta2024llama32} & 0.617 & 0.936 & 1.227 & 1.533 & 0.377 & 0.319 & 0.700\\
            Qwen3~\cite{yang2025qwen3} & 0.614 & 0.910 & 1.198 & 1.521 & 0.374 & 0.317 & 0.696 \\
            InternVL2~\cite{chen2024far} & \textbf{0.588} & \textbf{0.879} & \textbf{1.146} & \textbf{1.436} & \textbf{0.378} & \textbf{0.322} & \textbf{0.720} \\
            \bottomrule
        \end{tabular}
    \end{subtable}
    
    \vspace{2mm}
    
    \begin{subtable}{1.0\linewidth}
        \centering
        \caption{Comparison across model scales on InternVL2 series~\cite{chen2024far}.}
        \label{tab:scale-comparison}
        \scriptsize
        \setlength\tabcolsep{2.6mm}
        \begin{tabular}{lccccccc}
            \toprule
            \multirow{2.3}{*}{Model} & \multicolumn{4}{c}{Generation} & \multicolumn{3}{c}{Understanding}\\
            \cmidrule(lr){2-5}\cmidrule(lr){6-8}
            & 0s $\downarrow$ & 1s $\downarrow$ & 2s $\downarrow$ & 3s $\downarrow$ & M. $\uparrow$ & R. $\uparrow$ & C. $\uparrow$ \\
            \midrule
            0.8B & 0.601 & 0.890 & 1.150 & 1.434 & 0.376 & 0.319 & 0.708 \\
            1.8B & 0.588 & 0.879 & 1.146 & 1.436 & 0.378 & 0.322 & 0.720 \\
            3.8B & \textbf{0.557} & \textbf{0.768} & \textbf{0.991} & \textbf{1.255} & \textbf{0.383} & \textbf{0.325} & \textbf{0.742} \\
            \bottomrule
        \end{tabular}
    \end{subtable}
\end{table}

\section{Conclusion}
In this paper, we present \ours, a unified driving world model that integrates 3D scene understanding and future geometry prediction. By leveraging the BEV representation, we effectively consolidate multi-view visual information into a format compatible with LLMs. To facilitate interaction between semantic reasoning and geometric evolution, we introduce LLM-enhanced world queries to enable knowledge transfer. These queries then interact with LLM-encoded BEV features to generate latent representations for future timestamps via a Current-to-Future Link. To further ensure the structural consistency of predicted futures, we devise a Joint Geometric Optimization strategy that integrates explicit geometric constraints with implicit latent regularization to align internal representations with geometry-aware priors. Extensive evaluations validate the effectiveness of our approach. \ours~achieves strong performance, outperforming specialists in both generation and understanding tasks. We hope this work establishes a solid foundation for future research in interpretable and predictive driving systems.

\noindent\textbf{Limitation and future work.} While this work presents a solid exploration toward a unified driving world model, how to leverage the semantic priors encapsulated in pre-trained multi-modal large models for BEV input requires further investigation. Additionally, extending the generation to broader output modalities, such as future images and 3D occupancy, presents an interesting direction for comprehensive scene simulation.

\noindent\textbf{Acknowledgment.} This work was supported by the NSFC (62225603, 62441615, and 623B2038).

% \balance

{\small
\bibliographystyle{IEEEtran}
\bibliography{egbib}
}

\end{document}